\documentclass{article}

\usepackage[table,xcdraw,dvipsnames,svgnames,x11names]{xcolor}

\usepackage{arxiv}

\usepackage{times}
\usepackage{latexsym}

\usepackage[T1]{fontenc}

\usepackage[utf8]{inputenc}

\usepackage{microtype}

\usepackage{inconsolata}
\usepackage{amsmath}

\usepackage[skins]{tcolorbox} 
\tcbuselibrary{breakable} 
\newtcolorbox[auto counter, number within=section, list type=subsubsection, list inside=toc]{sectionbox}[2][]{
colback=white!98!gray, colframe=black, 
colbacktitle=white!90!gray, coltitle=black, 
fonttitle=\bfseries,
title={#2}, 
list entry={Comment \thetcbcounter\quad}
}

\definecolor{iccvblue}{rgb}{0.21,0.49,0.74}
\usepackage[pagebackref,breaklinks,colorlinks,allcolors=iccvblue]{hyperref}
\usepackage{pifont}    

\newcommand{\cmark}{\textcolor{green}{\ding{51}}}  
\newcommand{\xmark}{\textcolor{red}{\ding{55}}}

\definecolor{darkgreen}{RGB}{0,128,0}

\usepackage{listings}
\usepackage{wrapfig}
\usepackage{hyperref}       
\usepackage{url}            
\usepackage{booktabs}      
\usepackage{amsfonts}      
\usepackage{nicefrac}       
\usepackage{microtype}      
\usepackage{amsmath}
\usepackage{amssymb}
\usepackage[section]{placeins}
\usepackage{graphicx}
\usepackage{natbib}
\usepackage{booktabs}
\usepackage{tabularx}
\usepackage{subcaption}

\usepackage{adjustbox}
\usepackage{xspace}
\usepackage{colortbl} 
\usepackage{soul} 
\usepackage{tabularx}
\usepackage{stfloats}
\usepackage{algorithm,algorithmicx}
\usepackage[noend]{algpseudocode}
\usepackage{algcompatible}
\usepackage{pifont}
\usepackage{makecell}

\graphicspath{{./figures/}}
\usepackage{multirow}
\usepackage{enumitem}

\definecolor{lightgrey}{HTML}{dcdbdb}
\sethlcolor{lightgrey}
\definecolor{lightblue}{HTML}{E8F0FE}
\definecolor{lightblue}{HTML}{E8F0FE}
\definecolor{gray}{HTML}{9aa0a6}
\definecolor{lightpink}{HTML}{F48FB1}
\definecolor{lightred}{HTML}{FFCBC9}
\definecolor{lightcyan}{HTML}{80DEEA}
\definecolor{darkgreen}{rgb}{0.0, 0.5, 0.0}
\definecolor{babyblueeyes}{rgb}{0.1, 0.3, 0.5}
\definecolor{turquoisegreen}{rgb}{0.3, 0.6, 0.3}

\usepackage[skins]{tcolorbox} 
\tcbuselibrary{breakable}

\usepackage{lipsum}
\usepackage{soul}

\definecolor{greengrey}{rgb}{0.0, 0.3, 0.0} 
\colorlet{greengreywithhint}{greengrey!90!gray} 

\usepackage{multirow}
\usepackage{colortbl}
\usepackage{graphicx}
\usepackage{makecell} 
\usepackage{tcolorbox}

\author{Jipeng Zhang$^1$\footnotemark[1]\protect\phantom{\footnotesize 1},  Kehao Miao$^{2}$\thanks{\, Equal Contribution. Code are available at the following links: \url{https://github.com/2003pro/vlgenrm}.}\protect\phantom{\footnotesize 1},  \textbf{Renjie Pi}$^1$, \textbf{Zhaowei Wang}$^1$, \textbf{Runtao Liu}$^1$, \textbf{Rui Pan}$^3$, \textbf{Tong Zhang}$^3$ \\
\\
  $^1$The Hong Kong University of Science and Technology\\
  $^2$Nanyang Technological University, $^3$University of Illinois Urbana-Champaign \\
\texttt{\{jzhanggr,rpi,zwanggy,rliuay\}}@ust.hk\\ 
\texttt{kehao001@e.ntu.edu.sg}, \texttt{ruip4@illinois.edu,}
\texttt{tongzhang@tongzhang-ml.org}
}

\renewcommand{\shorttitle}{}

\newcommand{\method}{\textbf{VL-GenRM}}

\title{\method: Enhancing Vision-Language Verification via Vision Experts and Iterative Training}

\hypersetup{
pdftitle={A template for the arxiv style},
pdfsubject={q-bio.NC, q-bio.QM},
pdfauthor={David S.~Hippocampus, Elias D.~Striatum},
pdfkeywords={First keyword, Second keyword, More},
}

\begin{document}
\maketitle

\maketitle
\begin{abstract}
Reinforcement Fine-Tuning (RFT) with verifiable rewards has advanced large language models but remains underexplored for Vision-Language (VL) models. The Vision-Language Reward Model (VL-RM) is key to aligning VL models by providing structured feedback, yet training effective VL-RMs faces two major challenges. First, the bootstrapping dilemma arises as high-quality training data depends on already strong VL models, creating a cycle where self-generated supervision reinforces existing biases. Second, modality bias and negative example amplification occur when VL models hallucinate incorrect visual attributes, leading to flawed preference data that further misguides training. To address these issues, we propose an iterative training framework leveraging vision experts, Chain-of-Thought (CoT) rationales, and Margin-based Rejection Sampling. Our approach refines preference datasets, enhances structured critiques, and iteratively improves reasoning. Experiments across VL-RM benchmarks demonstrate superior performance in hallucination detection and multimodal reasoning, advancing VL model alignment with reinforcement learning.
\end{abstract}    
\section{Introduction}
\label{sec:intro}


Recent open-source work like DeepSeek R1~\citep{DeepSeek-R1} has highlighted Reinforcement Fine-Tuning (RFT) with verifiable rewards as a promising approach for improving large models such as OpenAI’s o1~\citep{OpenAI_O1}. While these techniques have shown success in text-based models, their application to Vision-Language (VL) models remains underexplored, despite the growing importance of VL reasoning models in multimodal AI. The Vision-Language Reward Model (VL-RM)~\citep{li2024vlrewardbench,Aligning-LLMs-with-RLHF, li2023silkie}, also referred to as a vision-language verifier, plays a crucial role in refining VL reasoning models~\citep{liu2025visual} by providing structured feedback to enhance response quality. The success of RFT in this domain depends on feedback quality, i.e., accuracy of VL-RM, highlighting the need for improved training strategies~\citep{bradley1952rank} of VL-RM.

\begin{wrapfigure}{r}{0.48\textwidth}
  \vspace{-4mm}
  \centering
  \includegraphics[width=0.95\linewidth]{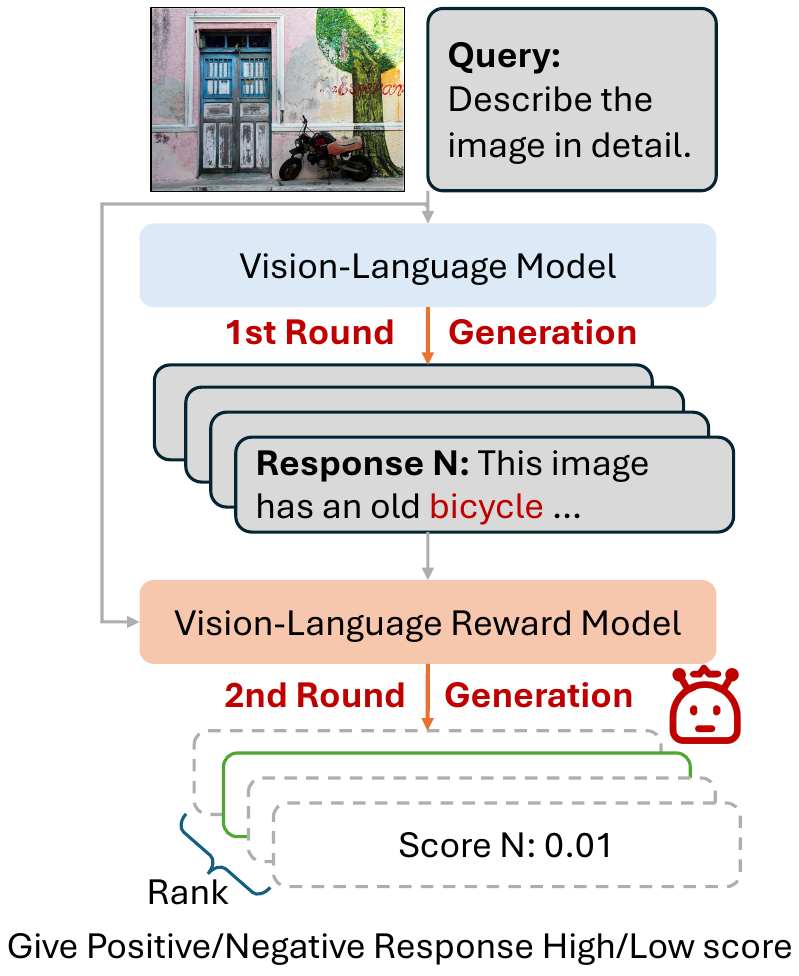}
   \caption{Illustration of the Vision-Language Reward Model (VLRM) pipeline, highlighting key challenges in training. \textbf{Issue 1 (Bootstrapping Pitfalls):} The VLM generates responses based on self-produced data, leading to potential hallucinations (e.g., the nonexistent \textit{"bicycle"}). While such errors can be mitigated using object detection expert models, relying solely on self-generated supervision risks reinforcing biases. \textbf{Issue 2 (Inherited Modality Bias \& Negative Example Amplification):} The VLRM evaluates these flawed responses, but biases from the first round persist, as shown by the \textit{confusing agent icon} in the second-generation step. Without structured reasoning or external supervision, these biases can be amplified rather than corrected, highlighting the need for improved alignment strategies.}
   \label{fig:VL-RM_example}
  \vspace{-6mm}
\end{wrapfigure}

As illustrated in Figure~\ref{fig:VL-RM_example} , we identified 2 primary challenges for the training of VLRMs: 

\textbf{Bootstrapping Pitfalls and the “Ouroboros-Like” Challenge.} To minimize the need for extensive manual annotation, most Vision-Language Model (VLM)~\citep{pi2024strengthening,li2023silkie,yu2024rlaif} and Reward Model (RM)~\citep{DPO, Self-Rewarding, Iterative,lee2023rlaif} training methods rely on larger, more powerful VLMs for bootstrapping—where stronger models generate or label data~\citep{Star, RAFT, RLHF-Workflow,chen2024self}. However, this creates a fundamental tail-eating-snake dilemma: high-quality data is essential for training strong VLMs, yet strong VLMs are needed to produce high-quality data. Breaking out of this “ouroboros-like” cycle requires introducing new expertise or external knowledge sources~\citep{pi2024image,chiu2025aide}, as relying solely on self-generated data risks reinforcing the model’s existing biases and limitations.

\textbf{Inherited Modality Bias in RM Training and Negative Example Amplification.} Negative responses are essential in RM training~\citep{bradley1952rank, zhang2024generative, zhong2024dpo,yang2024regularizing}, as they provide contrastive supervision~\citep{wang2024secrets} that helps refine a model’s evaluation capabilities. However, in VLRM training and inference, the inherent misalignment between text and images introduces compounding challenges~\citep{povid,pi2024strengthening}. The process is effectively multi-turn: first, the Vision-Language Model (VLM) generates a response, and then the VLRM evaluates it. Unfortunately, any cross-modal bias introduced in the first turn becomes “baked into” the negative examples used for direct evaluation in the second turn—potentially leading to inherited modality bias and negative transfer. For instance, VLMs frequently hallucinate nonexistent objects, misidentify attributes such as shape or color, or provide incorrect object counts. Ideally, these errors should be corrected by the VLM itself, yet the VLRM is still required to assess such flawed responses. Classical discriminative Reward Models~\citep{bradley1952rank} or direct generative RMs~\citep{zhang2024generative} typically rely on simple pairwise annotations (``Yes/No''), making them more prone to “negative example amplification.” A real-world analogy is seen in language learners: if their textbooks contain numerous incorrect grammar examples, they may internalize these errors rather than learn the correct forms, ultimately propagating misuses. Over multiple interactions, the need for chain-of-thought (CoT)~\citep{Chain-of-Thought-Prompting,pang2024iterative} rationales and accumulated context becomes critical in mitigating such biases.

Motivated by the challenges above, we introduce a novel training recipe to tackle these issues in training VLRMs. On the data side, we propose incorporating specialized visual knowledge~\citep{liu2023grounding,wu2019detectron2} and chain-of-thought rationales~\citep{zhang2024generative} to provide more constructive guidance.  On the training side, we employ preference-learning techniques~\citep{PPO,ouyang2022training} drawn from reinforcement learning (RL) in an iterative~\citep{RAFT,Star,touvron2023llama2} fashion—allowing the model’s generation to adapt toward more preferred outputs across multiple rounds. This RL-based method has already proven successful in large language models~\citep{2024gpt4o,bai2022training,gemini2023}, and one of our primary goals is to extend these techniques to address the unique challenge of aligning different modalities in a VLRM setting.

Thus, we make the following contributions: 
\begin{itemize}
    \item  \textbf{Automated Construction of Preference Datasets via Vision Experts. } We leverage vision experts to generate large-scale preference datasets, improving supervision quality in our \method.
    \item \textbf{CoT-Enhanced VL-GenRM Training.} We incorporate chain-of-thought rationale generation techniques to systematically guide VL-GenRM training. This structured reasoning process increases the effective correct description part in the dataset, mitigating the limitations of self-generated data and reinforcing coherent reward modeling.
    \item \textbf{ Iterative Bootstrapping with Margin-based Rejection Sampling.} We refine VL-GenRM’s reasoning through iterative fine-tuning on successful rationales, which are selected through the margin between reward signals of positive and negative examples.
    \item \textbf{Comprehensive Evaluation.} We validate our approach across VL-RM benchmarks and Best-of-N sampling, demonstrating improved multimodal reasoning and evaluation.
\end{itemize}

\section{Related Work}

\label{sec:related_work}
\noindent\textbf{Vision-Language Modeling.} Recent progress in vision-language models (VLMs) stems from integrating large language models (LLMs) with vision encoders via adaptation layers~\citep{liu2023llava,dai2023instructblip}. Key advancements focus on (1) curating high-quality multimodal datasets~\citep{zhang2024llavar,chen2023sharegpt4v}, (2) improving architectures for enhanced pixel-text reasoning~\citep{li2024llavaov}, and (3) optimizing training with reinforcement learning from human feedback (RLHF) to mitigate hallucinations~\citep{povid,pi2024strengthening}. However, robust reward modeling remains a challenge. We explore VLM-based reward training to enhance structured evaluation and reasoning.  

\noindent\textbf{Specialized Vision Expert Models.} Vision expert models specialize in object detection~\citep{girshick2015fast, yao2021g, carion2020end} and depth estimation~\citep{kim2022globallocal, yang2024depth}, enabling precise visual understanding. Recent work~\citep{gu2021open, liu2023grounding, yao2024detclipv3} shows their effectiveness in specialized tasks. We leverage vision experts to improve object-level verification, refining multimodal reasoning and reward modeling.

\noindent\textbf{Reward Models.} Reward models (RMs) are essential in reinforcement learning from human feedback (RLHF) and preference-based optimization~\citep{bradley1952rank,DPO}. Traditional RMs use binary classification or preference modeling to rank responses~\citep{ouyang2022training}, with early improvements focusing on better preference data and token-wise dense rewards~\citep{pi2024strengthening,lee2023rlaif,zhong2024dpo}. Recent work explores diverse RM types, such as outcome-based and process-based models~\citep{lightman2023let,zhang2024entropy,wang2023math}. Generative reward models (GenRMs)~\citep{zhang2024generative} leverage token probabilities instead of classification scores, aligning with LLMs' generative nature and enabling Chain-of-Thought (CoT) reasoning~\citep{Chain-of-Thought-Prompting}. Additionally, LLM-as-a-judge~\citep{zheng2023judging} eliminates separately trained RMs, while Direct Preference Optimization (DPO)~\citep{DPO} aligns models with human preferences without explicit rewards. Despite progress in text-based RMs, vision-language reward models (VL-RMs) remain underexplored~\citep{li2024vlrewardbench}, facing challenges in visual grounding, hallucination detection, and structured reasoning. Early efforts like VLFeedback~\citep{li2023silkie} and LLaVA-Critic~\citep{xiong2024llavacritic} introduce multimodal preference datasets and critique-based training. Our work advances this area by developing a generative VL-RM with iterative optimization, vision expert integration, and Best-of-N selection to improve multimodal reasoning consistency.

\noindent\textbf{Iterative RL.} Iterative reinforcement learning (RL) refines reward models through human-in-the-loop feedback. \textbf{Proximal Policy Optimization (PPO)}~\citep{PPO} is central to RLHF, iteratively improving response quality~\citep{ouyang2022training}. \textbf{Direct Preference Optimization (DPO)}~\citep{DPO} simplifies PPO-based RLHF by reformulating it as an offline optimization problem. Beyond PPO, \textbf{rejection sampling} methods like STaR~\citep{Star} and RAFT~\citep{RAFT} enhance preference learning by filtering suboptimal responses. Recently, \textbf{Iterative DPO}~\citep{Iterative,RLHF-Workflow} has gained traction, with variants like Pairwise Cringe Loss~\citep{xu2023some} and ReST~\citep{ReST} refining preference learning iteratively. SPIN~\citep{chen2024selfplay} extends DPO by integrating human-labeled winning responses, while \textbf{Self-Rewarding LLMs}~\citep{Self-Rewarding} generate preference pairs for better instruction following, though with limited reasoning gains. While widely applied to text-based models, iterative RL for \textbf{Vision-Language Models (VLMs)} remains underexplored. Our work pioneers \textbf{iterative RL in VL-RM training}, incorporating vision experts and multimodal reasoning enhancements to improve preference learning in multimodal contexts.

\section{Background}
\label{sec:background}

\begin{figure*}
  \centering
    \includegraphics[width=0.98\linewidth]{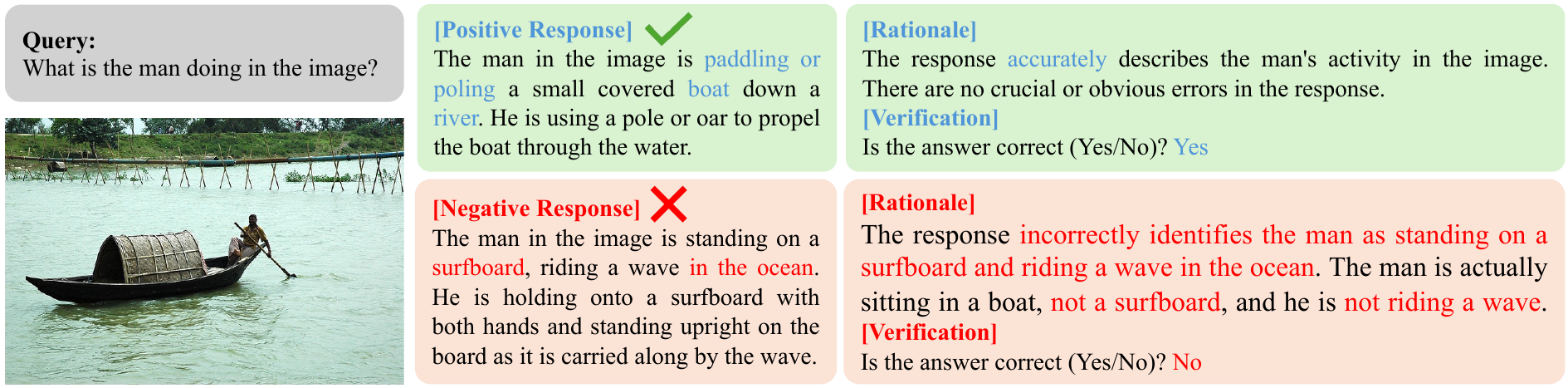}
    \caption{An example of a pairwise VLGenRM training data.}
    \label{fig:pair_data_example}
\end{figure*}

An autoregressive vision-language model generates an output sequence 
\( \mathbf{y} = (y_1, y_2, \dots, y_T) \) given an input image \( I \) 
and an input context \( \mathbf{x} \) 
(e.g., a textual description or question) by predicting tokens one at a time (i.e., next token prediction), based on the previously generated tokens. Assuming that the model is 
parameterized by \( \theta \), the conditional probability distribution 
of generating a sequence \( \mathbf{y} \) given \( I \) and \( \mathbf{x} \) is

\[
p_{\theta}(\mathbf{y} \mid I, \mathbf{x}) = \prod_{t=1}^{T} p_{\theta}(y_t \mid I, \mathbf{x}, \mathbf{y}_{<t})
\]

with \( \mathbf{y}_{<t} = (y_1, y_2, \dots, y_{t-1}) \). 
For ease of notation, we define \( p_{\theta}(y_t \mid I, \mathbf{x}) := p_{\theta}(y_t \mid I, \mathbf{x}, \mathbf{y}_{<t}) \). 


\noindent \textbf{Vision-Language Reward Modeling.} The vision-language (VL) reward model \( r_{\theta} \) assigns a score to a given input to assess the quality of the response \( y \):
\begin{equation}
    r_{\theta}(I, \mathbf{x}, y) = f_{\theta}(I, \mathbf{x}, y),
\end{equation}
where \( f_{\theta}(\cdot) \) is a learnable scoring function, typically implemented using a deep neural network. The training dataset \( \mathcal{D}_{\text{VL}} \) consists of tuples containing an image, a context, and both preferred and rejected responses:
\begin{equation}
    \mathcal{D}_{\text{VLRM}} = \{(I, \mathbf{x}, y^+, y^-) \},
\end{equation}
where:\( I \) is the input image,\( \mathbf{x} \) is the input context (e.g., a question or description),\( y^+ \) is the preferred response selected by humans, \( y^- \) is the less preferred or incorrect response.

Generally, reward modeling for vision-language models follows BT model, which aims to distinguish between the chosen response \( y^+ \) and the rejected response \( y^- \) 
given an image \( I \) and an input context \( \mathbf{x} \):
\begin{equation}
\begin{aligned}
    & \mathcal{L}_{\text{reward}}(\theta) = \\
    & -\mathbb{E}_{(I, \mathbf{x}, y^+, y^-) \sim \mathcal{D}} 
    \left[
    \log \sigma \Bigl( 
    r_{\theta}(I, \mathbf{x}, y^+) 
    - 
    r_{\theta}(I, \mathbf{x}, y^-)
    \Bigr)
    \right],
\end{aligned}
\end{equation}
where \( r_{\theta}(I, \mathbf{x}, y) \) is the reward score for image \( I \), context \( \mathbf{x} \), 
and response \( y \). \( \sigma(\cdot) \) is the sigmoid function. 
Minimizing this loss ensures higher scores for human-preferred responses, 
enabling the trained reward model to guide vision-language model optimization.

\noindent \textbf{Generative Reward Modeling (GenRM).}  
GenRM formulates verification as a token prediction task, where a VLM learns to predict correctness labels given an image \( I \), input context \( \mathbf{x} \), and response \( y \). The training dataset consists of labeled problem-solution pairs:

\[
\mathcal{D}_{\text{GenRM}} = \{ (\mathbf{x}, y^+, I), p, \text{‘Yes’} \} \cup \{ (\mathbf{x}, y^-, I), p, \text{‘No’} \},
\]

where \( p \) is a fixed prompt (``Is the most recent final answer correct (Yes or No)'') that instructs the model to verify the correctness of \( y \).  
At inference, correctness likelihood is used as the model’s confidence score:

\[
r_{\text{GenRM}}(\mathbf{x}, y, I) = p_{\theta}(\text{‘Yes’} \mid \mathbf{x}, y, I, p).
\]

This approach enables direct verification through token probabilities via instruction tuning. The training objective of GenRM is Supervised Fine-Tuning (SFT) loss:  $\mathcal{L}_{\text{GenRM}} = \mathcal{L}_{\text{SFT}}$.


For the formulation of SFT, DPO, VLM-as-a-Judge and BoN, please refer to the Appendix~\ref{appendix:formulation}.

\section{Data Collection}

\label{sec:data}

\begin{wrapfigure}{r}{0.48\textwidth}
  \vspace{-4mm}
  \centering
  \includegraphics[width=0.95\linewidth]{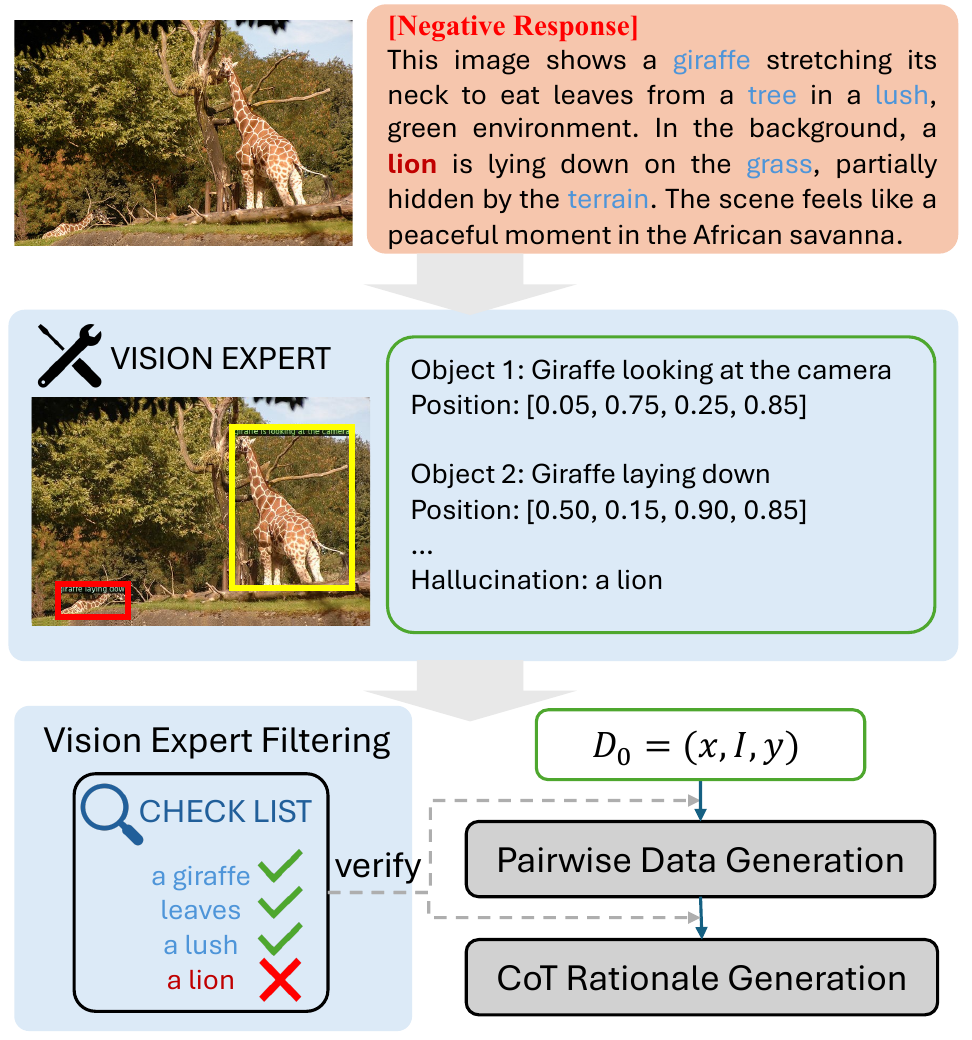}
    \caption{The pipeline starts with a hallucinated negative response, like misidentifying a "lion" in an image. A vision expert verifies objects, filters errors, refines the response, and generates rationales to enhance training data.}
    \label{fig:data_generation}
  \vspace{-6mm}
\end{wrapfigure}

In this section, we describe our two-phase data preparation approach for \method: Pairwise Data Generation and Chain-of-Thought (CoT) Generation and Verification. To ensure high-quality training data, we incorporate vision experts for object detection and verification, refining rejected responses to enhance preference datasets (Figure~\ref{fig:data_generation}).

\subsection{Pairwise Data Generation}

The data generation framework for VL-GenRM follows a structured three-step process to ensure high-quality negative responses for training. Given the original dataset $\mathcal{D}_{0} = \{(I, X, y)\}$, our goal is to construct a dataset \(\mathcal{D}_{pair}\) that consists of image-query-response pairs with both accurate positive responses and refined negative responses:

\begin{equation}
    \mathcal{D}_{pair} = \{(I, X, Y^+, Y^-_{\text{new}}, \hat{y})\}
\end{equation}

where \(I\) is the image, \(X\) is the input query, \(Y^+\) is the chosen response, \(Y^-_{\text{new}}\) is the refined negative response, and \(\hat{y} \in \{\text{Yes}, \text{No}\}\) is the binary label indicating correctness.

To achieve this, we follow three key steps: (1) \textbf{Negative Response Collection}, where a weak VLM generates incorrect responses; (2) \textbf{Vision Expert Filtering}, which verifies hallucinated responses by checking object presence in images; and (3) \textbf{Refinement and Augmentation}, where false rejections are corrected, and negative responses are modified to improve response diversity. This pipeline ensures that pairwise samples are \textbf{realistic, visually grounded, and semantically refined}, enhancing VL-GenRM’s reward modeling capabilities.

\subsubsection{Negative Response Collection}

In the \textbf{Negative Response Collection} phase, we construct an initial dataset \(\mathcal{D}_0 = \{(I, X, Y^+)\}\), where \(I\) is an image, \(X\) is the query, and \(Y^+\) is a chosen response. To introduce contrastive supervision, we generate negative responses using a weak vision-language model (VLM), producing:
$
Y^- = f_{\text{weak-VLM}}(I, X)
$
These responses serve as \textbf{plausible but incorrect} samples, mimicking common errors made by weaker models. However, some responses may contain near-correct answers, requiring additional filtering.

\subsubsection{Vision Expert Filtering}

In the \textbf{Vision Expert Filtering} phase, we eliminate hallucinated negatives by verifying object consistency. First, we extract mentioned objects from the generated negative response:$
\mathcal{O}(Y^-) = f_{\text{VLM}}(Y^-).
$Then, we detect actual objects in the image using an object detector (OD):$
\mathcal{O}^*(I) = f_{\text{OD}}(I).
$If \(\mathcal{O}(Y^-) \not\subseteq \mathcal{O}^*(I)\), the response is labeled as a hallucination and retained. For the remaining responses in the dataset, we will perform Refinement and Augmentation.

\subsubsection{Refinement and Augmentation}

In the \textbf{Refinement and Augmentation} phase, we correct false rejections and generate modified negatives to improve diversity. First, we determine if a rejected response is approximately correct by extracting objects from the image:
$
\mathcal{O}_I = g_{\text{OD}}(I).
$
A stronger VLM evaluates if \(Y^-\) is a valid response:

\begin{equation}
    \hat{y} = f_{\text{VLM}}(I, X, Y^-, \mathcal{O}_I), \quad \hat{y} \in \{\text{Yes}, \text{No}\}
\end{equation}

If correct, \(Y^-\) is flagged for replacement instead of rejection. Next, to diversify negative responses, we modify object mentions in \(Y^+\) by selecting two objects:$
\mathcal{O}_{\text{sampled}}(Y^+) \subseteq \mathcal{O}(Y^+)
.$ A new negative response is then generated by altering these objects:$
Y^-_{\text{new}} = f_{\text{VLM}}(\mathcal{O}_{\text{sampled}}, Y^+)
$

The final dataset \(\mathcal{D}_{pair} = \{(I, X, Y^+, Y^-_{\text{new}}, \hat{y})\}\) ensures that negative responses remain semantically valid yet distinct, strengthening the contrastive learning signal in VL-GenRM.

\subsection{Chain-of-Thought (CoT) Rationale Generation}

Given the dataset \(\mathcal{D}_{pair} = \{(I, X, Y^+, Y^-_{\text{new}}, \hat{y})\}\) from the previous stage, the goal of the CoT rationale generation process is to construct a dataset \(\mathcal{D}_{\text{train}}\) that provides structured reasoning rationales alongside response pairs, enabling VL-GenRM to better assess correctness and improve interpretability. The final dataset is formulated as:

\begin{equation}
    \mathcal{D}_{\text{train}} = \{(I, X, Y, c, \hat{y})\}
\end{equation}



where \(I\) is the image, \(X\) is the query, \(Y = \{Y^+, Y^-\}\), \(Y^+\) and \(Y^-\) are the chosen and rejected responses, respectively, \(c = \{c^+, c^-\}\), \(c^+\) and \(c^-\) are the CoT rationales explaining why each response is correct or incorrect, and \(\hat{y} \in \{\text{Yes}, \text{No}\}\) is the binary label.

To generate CoT rationales, we use a strong vision-language model (VLM) to produce step-by-step reasoning explanations for both responses:

\begin{equation}
    c^+ = f_{\text{VLM}}(I, X, Y^+, \mathcal{O}_I),\quad
    c^- = f_{\text{VLM}}(I, X, Y^-, \mathcal{O}_I)
\end{equation}

where \(\mathcal{O}_I\) represents detected objects in the image. These rationales help the model learn to justify its reward assignment, improving consistency in evaluating correctness.

To enhance data quality, we apply \textbf{selective rationale filtering}, prompting the model to focus only on missing key objects and critical errors, thereby reducing unnecessary hallucinations. Additionally, we introduce \textbf{external dataset augmentations}, generating multiple responses per question using a smaller VLM to increase reasoning diversity. If no incorrect responses are naturally found, we inject \textbf{random incorrect answers} to maintain a balanced dataset.





\section{Training Framework}
\label{sec:training}

\begin{figure*}
  \centering
    \includegraphics[width=0.98\linewidth]{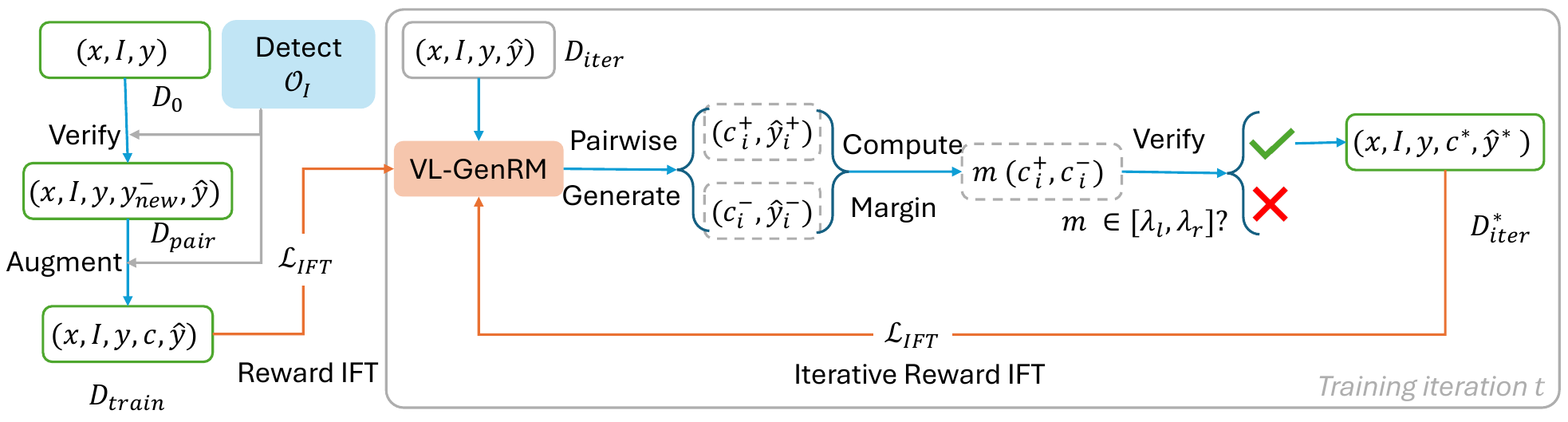}
    \caption{Iterative Training Pipeline of VL-GenRM. The training consists of two stages: \textbf{(1) Reward Initialization}—Given raw data \((x, I, y)\), OpenAI’s \(\mathcal{O}_I\) model detects correctness and generates refined annotations \((x, I, y_{\text{new}}, \hat{y})\). Contrastive pairs are constructed to form the structured dataset \(D_{\text{train}}\), which is used for instruction-following fine-tuning (\(\mathcal{L}_{\text{IFT}}\)). \textbf{(2) Iterative Refinement}—VL-GenRM simultaneously generates candidate rationales for both positive and negative responses, which are verified against reference outputs. A \textit{Margin-based Rejection Sampling} strategy filters the most informative rationales \((c^*, \hat{y}^*)\), refining the dataset \(D_{\text{iter}}\) for continued fine-tuning. This iterative approach enhances reward alignment, mitigates hallucinations, and improves multimodal reasoning performance.}
    \label{fig:training_pipeline}
\end{figure*}

As shown in Figure~\ref{fig:training_pipeline}, following standard post-training practices for large models, we adopt a two-stage training framework to optimize VL-GenRM: (1) \textbf{Rewarding instruction-following fine-tuning}, where the model learns structured reward modeling from CoT rationales, and (2) \textbf{Iterative optimization}, where the model refines itself via self-generated reasoning and reward alignment.

\subsection{Rewarding Instruction-Following Fine-Tuning}

To develop an initial reward model, we train VL-GenRM on a structured dataset containing both positive and negative reasoning responses with corresponding CoT rationales. The training dataset is $\mathcal{D}_{\text{train}} = \{(I, X, Y, c, \hat{y})\}$.

\paragraph{Rewarding instruction following learning (IFT).}  
We structure this training as an instruction-following task, where the model is required to assess a response and generate a reasoning-based evaluation. Specifically, we train the model to first generate a CoT rationale and then output a structured binary evaluation. From $\mathcal{D}_{\text{train}}$, We extract all the problems-solution pairs with correctness tokens as $\mathcal{D}_{\text{correct}} = \{(I, X, Y^+, c^+, \hat{y})\}$. Following~\citep{zhang2024generative}The training objective of this IFT stage is: 

\begin{equation}
    \mathcal{L}_{\text{IFT}}(\theta, \mathcal{D}_{\text{train}}) = 
    \mathcal{L}_{\text{SFT}}(\theta, \mathcal{D}_{\text{train}}) + 
    \lambda \mathcal{L}_{\text{SFT}}(\theta, \mathcal{D}_{\text{correct}}),
\end{equation}

where \( \lambda > 0 \) is a hyperparameter that controls the mixture ratio between verification (\(\mathcal{D}_{\text{train}}\)) and generating correct solutions (\(\mathcal{D}_{\text{correct}}\)). This unified training can improve both verification and generation performance. This step produces a refined dataset \(\mathcal{D}_{\text{train}}\), allowing VL-GenRM to develop structured reward alignment.

\subsection{Iterative Optimization}

Once an initial \method ~is trained, we apply an iterative optimization strategy to further refine the model’s ability to assess correctness and reasoning consistency. This process involves self-generated CoT rationales, Margin-based Rejection Sampling, and IFT. The iterative training dataset is defined as:

\begin{equation}
    \mathcal{D}_{\text{iter}} = \{(I, X, Y^+, Y^-, c^*_+, c^*_-)\}
\end{equation}

where \(c^*_+\) and \(c^*_-\) are the most informative rationales selected from multiple generated rationales. At the $t$-th iteration, the model undergoes the following steps in each iteration:

\begin{enumerate}
    \item \textbf{Generating CoT Rationales:}  
    The model simultaneously generates the reasoning rationales for both positive and negative response:
    \begin{equation}
        c^+ = f_{\text{RM}}(I, X, Y^+), \quad
        c^- = f_{\text{RM}}(I, X, Y^-)
    \end{equation}
    
    \item \textbf{Margin-based Rejection Sampling:}  
    We select the most informative rationales for both positive and negative responses using a margin-based scoring function~\citep{touvron2023llama2}:
    \begin{equation}
        m (c^+_i,c^-_i) = \text{score}(c^+_i) - \text{score}(c^-_i),
    \end{equation}
    
    \item \textbf{Data Augmentation and Filtering:} 
    We set a margin threshold range $[\lambda_l,\lambda_r]$ here. The rationale pairs with margin score $m (c^+_i,c^-_i)$ fall within this range are kept, yielding a progressively refined dataset $\mathcal{D}_{\text{iter}}^{*}$.
\end{enumerate}

To integrate these self-improving reasoning trajectories into VL-GenRM, we apply lightweight fine-tuning using Low-Rank Adaptation (LoRA)~\citep{hu2022lora}.
By leveraging self-correcting rejection sampling and iterative refinement, our approach enhances the model’s ability to perform structured reward learning, aligning CoT reasoning with robust verification capabilities.

\section{Implementation and Evaluation Settings}

\label{sec:implementation}

\subsection{Implementation}


For detailed implementation, training configurations and inference hyperparameters, please refer to Appendix~\ref{appendix:implementation}.

\subsection{Evaluation Metrics and Benchmarks}

To fully validate the effectiveness of our method, we employ two evaluation strategies: (1) evaluating directly on VLRewardBench following~\citep{li2024vlrewardbench} and (2) using the trained VL-GenRM as a test-time verifier on LLaVA-Wild~\citep{liu2023llava} to assess Best-of-N accuracy.

\subsubsection{RewardBench}
We evaluate VL-GenRM following~\citep{li2024vlrewardbench}, assessing alignment, reasoning, and multimodal understanding.  

\noindent \textbf{General Instruction Following.} We sample 183 instances from WildVision~\citep{lu2024wildvision} and VLFeedback~\citep{li2023silkie}. WildVision covers diverse multimodal queries with human-verified annotations, while VLFeedback uses GPT-4V-based assessments with high-quality preference labels.  

\noindent \textbf{Hallucination.} For visual hallucination detection, we sample 749 examples from POVID~\citep{povid}, RLAIF-V~\citep{yu2024rlaif}, and RLHF-V~\citep{yu2024rlhfv}. POVID introduces controlled noise for robustness testing, RLAIF-V employs automated verification, and RLHF-V refines labels via human preference lables.  

\noindent \textbf{Reasoning.} We evaluate complex multimodal reasoning using 318 pairs from MMMU-Pro~\citep{yue2024mmmu} and MathVerse~\citep{zhang2024mathverse}. MMMU-Pro assesses high-level multimodal inference across disciplines, while MathVerse focuses on visual mathematical reasoning tasks.

\subsubsection{Test-time Best-of-N accuracy}
To further assess the effectiveness of VL-GenRM, we evaluate its ability to serve as a test-time verifier by measuring \textbf{Best-of-N accuracy}. This approach follows prior works~\citep{yang2024regularizing,liang2024improving,hosseini2024v}, where we sample multiple candidate responses, rank them using the trained reward model, and select the highest-scoring response as the final answer. Here, we select Qwen-2-7B~\citep{wang2024qwen2}, InterVL-4B~\citep{chen2024expanding} and LLaVA-Next-8B~\citep{liu2024llavanext} to generate N responses and then employ the VLRM to select the BoN. 

Formally, given an input \((I, X)\), we generate a set of \(N\) candidate responses:
$
\mathcal{Y} = \{Y_1, Y_2, ..., Y_N\} \sim P(Y|I, X)
$
where each \(Y_i\) is sampled from the model’s response distribution. The reward model assigns a score \(s_i\) to each response:
$
s_i = f_{\text{RM}}(I, X, Y_i)
$
We then select the response with the highest score:
$
Y^* = \arg\max_{Y_i \in \mathcal{Y}} s_i
$
We evaluate the effectiveness of test-time BoN on the popular
LLaVA-Wild~\citep{liu2023llava} benchmark.

\subsection{Baseline Models and Methods}

\textbf{Evaluated Models.} We evaluate four Proprietary VLMs, GPT-4o~\citep{2024gpt4o}, Gemini-1.5-Pro~\citep{gemini2023}, Claudge-3.5-Sonnet~\citep{bai2022training} and Qwen-VL-Max~\citep{bai2023qwenvl}. For the open-source VLM, we include VLMs with parameters ranging from 7B to 90B: Llama-3.2~\citep{dubey2024llama}, Molmo~\citep{deitke2024molmo}, DeepSeek-VL~\citep{lu2024deepseek}, Aria~\citep{li2024aria}, MammoTH-VL~\citep{guo2024mammoth}, Qwen2-VL~\citep{wang2024qwen2}. It is worth noting that Aria and DeepSeek own Mixture-of-Expert LLM. 

\noindent\textbf{Evaluated Methods.} We compare our approach against four vision-language reward modeling methods.  
\textbf{BT-RM (Bradley-Terry Reward Model)} optimizes a reward function via pairwise ranking, distinguishing preferred and rejected responses based on the Bradley-Terry model.  
\textbf{VLM-as-a-Judge} employs a vision-language model (VLM) to directly assess response quality, optionally comparing generated responses to reference answers.  
\textbf{DPO (Direct Preference Optimization)} reformulates preference learning as direct policy optimization, aligning response probabilities with human preferences without explicit reward modeling.  
\textbf{Direct GenRM (Direct Generative Reward Modeling)} trains a VLM to classify responses as correct or incorrect using token prediction, with correctness likelihood serving as the reward score.


    
    
    

\section{Experiment}

\label{sec:experiment}

\subsection{Comparison with other VLMs.}

\begin{wraptable}{r}{0.45\textwidth}
    \vspace{-6mm}
    \centering
    \small  
    \renewcommand{\arraystretch}{1.1}
    \begin{tabular}{l c c c c}
        \toprule
        \textbf{Model} & \textbf{Gen.} & \textbf{Hall.} & \textbf{Reason.} & \textbf{Overall} \\
        \midrule
        \multicolumn{5}{l}{\textit{Proprietary VLMs}} \\
        GPT-4o & 49.1 & 67.6 & \textbf{70.5} & 65.8 \\
        Gemini & 50.8 & 72.5 & 64.2 & 67.2 \\
        Claude & 43.4 & 55.0 & 62.3 & 55.3 \\
        Qwen-VL & 40.6 & 46.0 & 57.6 & 48.2 \\
        \midrule
        \multicolumn{5}{l}{\textit{Open-Source VLMs}} \\
        Llama & 42.6 & 57.3 & 61.7 & 56.2 \\
        Molmo & 33.9 & 42.3 & 54.9 & 44.1 \\
        DeepSeek & 29.7 & 23.8 & 50.9 & 30.3 \\
        Aria & 37.9 & 33.1 & 64.0 & 41.1 \\
        Pixtral & 35.6 & 25.9 & 59.9 & 35.8 \\
        MAmmoTH & 42.0 & 41.0 & 53.0 & 45.2 \\
        Qwen2 & 31.6 & 19.1 & 51.1 & 28.3 \\
        \midrule
        \rowcolor{gray!20}
        VL-GenRM & \textbf{54.6} & \textbf{82.4} & 58.8 & \textbf{72.3} \\
        \bottomrule
    \end{tabular}
    \caption{Comparison of Proprietary and Open-Source VLMs.}
    \label{tab:vlm_main_comparison}
    \vspace{-14mm}
\end{wraptable}

As shown in Table~\ref{tab:vlm_main_comparison}, \textbf{VL-GenRM (7B)} achieves state-of-the-art performance among open-source VLMs in general QA and hallucination robustness, outperforming much larger models like LLaMA-3.2 (90B) and Molmo (72B). This highlights that a well-designed \textbf{7B model can surpass significantly larger counterparts} in these areas.

However, VL-GenRM lags in reasoning due to its \textbf{object detection-based vision module}, which improves general understanding and hallucination resistance but is less suited for abstract reasoning requiring fine-grained scene analysis.

Overall, VL-GenRM demonstrates \textbf{strong generalization and hallucination robustness despite its compact size}, with room for improvement in reasoning. Full results can be found in Appendix~\ref{appendix:full_vlm_comparison}.




\subsection{Comparison with other training methods.}
As shown in Table~\ref{tab:vlm_main_comparison}, \textbf{VL-GenRM (7B)} achieves state-of-the-art performance among open-source VLMs in general QA and hallucination robustness, outperforming much larger models like LLaMA-3.2 (90B) and Molmo (72B). This highlights that a well-designed \textbf{7B model can surpass significantly larger counterparts} in these areas.

However, VL-GenRM lags in reasoning due to its \textbf{object detection-based vision module}, which improves general understanding and hallucination resistance but is less suited for abstract reasoning requiring fine-grained scene analysis.

Overall, VL-GenRM demonstrates \textbf{strong generalization and hallucination robustness despite its compact size}, with room for improvement in reasoning through enhanced multi-modal fusion.

\begin{table*}[h]
    \centering
    \begin{tabular}{l cc c c c c}
        \toprule
        \textbf{Model} & \textbf{Generative} & \textbf{Test Time} & \textbf{General} & \textbf{Hallucination} & \textbf{Reasoning} & \textbf{Overall} \\
        \midrule
        \textit{Base: Qwen-VL-7B} & & & & & & \\
        Bradley-Terry RM & \xmark & \xmark & 47.5 & 47.8 & 53.5 & 49.2 \\
        VLM-as-a-Judge & \xmark & \xmark & 54.9 & 53.6 & \textbf{69.8} & 57.9 \\
        DPO & \xmark & \xmark & 53.6 & 70.1 & 46.9 & 61.8 \\
        Direct GenRM & \cmark & \xmark & \textbf{63.9} & 68.9 & 54.4 & 64.5 \\
        \rowcolor{gray!20}
        \method & \cmark & \cmark & 54.6 & \textbf{82.4} & 58.8 & \textbf{72.3} \\
        \midrule
        \textit{Base: InternVL-4B} & & & & & & \\
        Direct GenRM & \cmark & \xmark & 53.0 & 76.6 & 56.0 & 67.9 \\
        \rowcolor{gray!20}
        \method & \cmark & \cmark & 55.7 & 80.4 & 59.4 & 71.4 \\
        \bottomrule
    \end{tabular}
    \caption{Comparison of different training methods. We categorize reward models (RMs) based on whether they are \textbf{generative} (capable of directly generating reward scores from ``Yes/No'' token.) and whether they introduce additional \textbf{test-time computation (Test Time)} overhead. Generative RMs are preferable for better inference performance. Additionally, models that use extra test-time computation are more capable.
Among the evaluated methods, \textbf{VL-GenRM} consistently outperforms others and achieves the best overall performance.}
    \label{tab:vlm_training_comparison}
\end{table*}

\begin{wraptable}{r}{0.55\textwidth}
\centering
    \begin{tabular}{l c c c c}
        \toprule
        \textbf{Model} & \textbf{BT } & \textbf{ GenRM} & \textbf{ IFT} & \textbf{Iteration} \\
        \midrule
        Qwen2.5-VL-7B & 74.2 & 74.2 & 73.5 & 73.0 \\
        $\Delta$ & \textcolor{darkgreen}{$\uparrow$ 2} & \textcolor{darkgreen}{$\uparrow$ 2} & \textcolor{darkgreen}{$\uparrow$ 1.3} & \textcolor{darkgreen}{$\uparrow$ 0.8} \\
        \midrule
        InternVL2.5-VL-4B & 75.5 & 75.2 & 78.3 & 78.7 \\
        $\Delta$ & \textcolor{red}{$\downarrow$ 2.7} & \textcolor{red}{$\downarrow$ 3.0} & \textcolor{darkgreen}{$\uparrow$ 0.1} & \textcolor{darkgreen}{$\uparrow$ 0.5} \\
        \midrule
        LLaVA-Next-8B & 73.3 & 76.8 & 78.3 & 79.5 \\
        $\Delta$ & \textcolor{red}{$\downarrow$ 4.7} & \textcolor{red}{$\downarrow$ 1.2} & \textcolor{darkgreen}{$\uparrow$ 0.3} & \textcolor{darkgreen}{$\uparrow$ 1.5} \\
        \midrule
        Average & \textcolor{red}{$\downarrow$ 1.8} & \textcolor{red}{$\downarrow$ 0.7} & \textcolor{darkgreen}{$\uparrow$ 0.6} & \textbf{\textcolor{darkgreen}{$\uparrow$ 0.9}} \\
        \bottomrule
    \end{tabular}
    \caption{Performance improvement brought by the proposed training pipeline. Green arrow denotes the percentage improvement over the baseline without Best-of-N sampling (N=16).}
    \label{tab:best_of_n_comparison}
\end{wraptable}


\subsection{Test-time Best-of-N Evaluation}

Table~\ref{tab:best_of_n_comparison} evaluates \textbf{VL-GenRM} under \textbf{Best-of-N (BoN) accuracy}, which measures its ability to act as a \textbf{test-time verifier} by selecting the best response among multiple candidates. This provides a more \textbf{robust} validation compared to single-response reward alignment.

Across models like \textbf{Qwen2.5-VL-7B, InternVL2.5-VL-4B, and LLaVA-Next-8B}, VL-GenRM consistently outperforms \textbf{BT RM} and \textbf{Direct GenRM}, excelling in both reward modeling and test-time verification. \textbf{Reward IFT + Iteration} further enhances performance, indicating that iterative refinement improves reward alignment. Additionally, results confirm that \textbf{VL-GenRM is model-agnostic}, demonstrating adaptability across architectures.

Overall, these findings highlight \textbf{VL-GenRM} as a \textbf{scalable and effective} reward modeling solution, improving generation quality while maintaining efficiency.




\subsection{Ablation}
\begin{table*}[ht]
\centering
\begin{minipage}[t]{0.48\textwidth}
    \centering
    \small
    \resizebox{\textwidth}{!}{
    \begin{tabular}{l c c c c}
        \toprule
        Model & General & Hallu. & Reasoning & Overall \\
        \midrule
        \textbf{Qwen-VL-7B} & & & & \\
        + pair data            & 53.6 & 70.1 & 46.9 & 61.8 \\
        + verified pair        & 51.4 & 78.2 & 50.6 & 67.3 \\
        + descriptive CoT pair & 44.8 & 60.4 & 56.0 & 57.0 \\
        + critique CoT pair    & 54.6 & 78.5 & 62.3 & 70.9 \\
        \midrule
        \textbf{InternVL-4B} & & & & \\
        + pair data            & 51.4 & 77.2 & 53.1 & 67.3 \\
        + verified pair        & 53.0 & 76.6 & 56.0 & 67.9 \\
        + descriptive CoT pair & 48.1 & 67.4 & 58.8 & 62.4 \\
        + critique CoT pair    & 55.7 & 80.4 & 59.4 & 71.4 \\
        \bottomrule
    \end{tabular}}
    \caption{Performance comparison of different data augmentation strategies.}
    \label{tab:pair_data_comparison}
\end{minipage}
\hfill
\begin{minipage}[t]{0.48\textwidth}
    \centering
    \small
    \vspace{-2cm}
    \resizebox{\textwidth}{!}{
    \begin{tabular}{lcccc}
        \toprule
        Model & General & Hallu. & Reasoning & Overall \\
        \midrule
        Baseline & 51.4 & 78.2 & 50.6 & 67.3 \\
        Reward IFT & 54.6 & 78.5 & 62.3 & 70.9 \\
        Iteration 1 & 55.2 & 82.1 & 58.8 & 72.2 \\
        Iteration 2 & 53.0 & 83.6 & 57.9 & 72.6 \\
        \bottomrule
    \end{tabular}}
    \caption{Performance of different training steps based on QwenVL-7B.}
    \label{tab:iteration_comparison}
\end{minipage}
\end{table*}

\textbf{Data Ablation.} Table~\ref{tab:pair_data_comparison} compares different data augmentation strategies. \textbf{``+ pair data''} is the baseline without \textbf{CoT rationales} or \textbf{vision expert verification}, leading to the weakest performance. \textbf{``+ verified pair''} improves hallucination robustness but lacks \textbf{test-time computation}, limiting reasoning gains. We further explore \textbf{CoT rationale generation}. \textbf{``+ descriptive CoT pair''} fails due to inherited modality bias and negative example amplification. In contrast, \textbf{``+ critique CoT pair''} enables effective test-time computation, improving both reasoning and hallucination control. This validates that \textbf{critique-based CoT} is essential for reasoning-aware supervision. Readers can refer to Table~\ref{tab:critique_prompt} and Table~\ref{tab:descriptive_prompt} for the detailed prompt used for generating such critique/descriptive CoT. We also made a data contamination analysis in Appendix~\ref{appendix:contamination}. The results show that the improvement is not merely due to including data from the same distribution.

\noindent\textbf{Training Method Ablation.} Table~\ref{tab:iteration_comparison} validates the effectiveness of our training design. \textbf{``Reward IFT''} significantly boosts reasoning and overall performance , while \textbf{``Iteration 1''} further enhances hallucination robustness. \textbf{``Iteration 2''} shows marginal gains, indicating saturation in our current setup. However, we expect further iterations to remain beneficial with \textbf{larger models and datasets}, highlighting the scalability of our approach.

    

\section{Conclusion}

\label{sec:conclusion}

In this work, we propose an iterative optimization framework for VL-GenRMs, incorporating vision experts, Chain-of-Thought (CoT) reasoning, and Margin-based Rejection Sampling to improve multimodal reward modeling. By leveraging vision experts for automated preference data construction and using CoT-based criticalization, we enhance data quality and mitigate bootstrapping challenges. Our iterative fine-tuning strategy allows the model to self-generate and refine rationales, leading to more consistent reasoning and better alignment. Experimental results demonstrate significant improvements in VL-RM performance and VLLM training, providing a scalable and effective solution for multimodal reinforcement learning.

\bibliographystyle{unsrtnat}
\bibliography{template}

\begin{thebibliography}{71}
\providecommand{\natexlab}[1]{#1}
\providecommand{\url}[1]{\texttt{#1}}
\expandafter\ifx\csname urlstyle\endcsname\relax
  \providecommand{\doi}[1]{doi: #1}\else
  \providecommand{\doi}{doi: \begingroup \urlstyle{rm}\Url}\fi

\bibitem[Guo et~al.(2025)Guo, Yang, Zhang, Song, Zhang, Xu, Zhu, Ma, Wang, Bi, et~al.]{DeepSeek-R1}
Daya Guo, Dejian Yang, Haowei Zhang, Junxiao Song, Ruoyu Zhang, Runxin Xu, Qihao Zhu, Shirong Ma, Peiyi Wang, Xiao Bi, et~al.
\newblock Deepseek-r1: Incentivizing reasoning capability in llms via reinforcement learning.
\newblock \emph{arXiv preprint arXiv:2501.12948}, 2025.

\bibitem[Jaech et~al.(2024)Jaech, Kalai, Lerer, Richardson, El-Kishky, Low, Helyar, Madry, Beutel, Carney, et~al.]{OpenAI_O1}
Aaron Jaech, Adam Kalai, Adam Lerer, Adam Richardson, Ahmed El-Kishky, Aiden Low, Alec Helyar, Aleksander Madry, Alex Beutel, Alex Carney, et~al.
\newblock Openai o1 system card.
\newblock \emph{arXiv:2412.16720}, 2024.

\bibitem[Li et~al.(2024{\natexlab{a}})Li, Wei, Xie, Yang, Song, Wang, An, Liu, Li, Lin, et~al.]{li2024vlrewardbench}
Lei Li, Yuancheng Wei, Zhihui Xie, Xuqing Yang, Yifan Song, Peiyi Wang, Chenxin An, Tianyu Liu, Sujian Li, Bill~Yuchen Lin, et~al.
\newblock Vlrewardbench: A challenging benchmark for vision-language generative reward models.
\newblock \emph{arXiv preprint arXiv:2411.17451}, 2024{\natexlab{a}}.

\bibitem[Sun et~al.(2024)Sun, Shen, Cao, Liu, Li, Shen, Gan, Gui, Wang, Yang, et~al.]{Aligning-LLMs-with-RLHF}
Zhiqing Sun, Sheng Shen, Shengcao Cao, Haotian Liu, Chunyuan Li, Yikang Shen, Chuang Gan, Liang-Yan Gui, Yu-Xiong Wang, Yiming Yang, et~al.
\newblock Aligning large multimodal models with factually augmented rlhf.
\newblock In \emph{ACL}, 2024.

\bibitem[Li et~al.(2023)Li, Xie, Li, Chen, Wang, Chen, Yang, Wang, and Kong]{li2023silkie}
Lei Li, Zhihui Xie, Mukai Li, Shunian Chen, Peiyi Wang, Liang Chen, Yazheng Yang, Benyou Wang, and Lingpeng Kong.
\newblock Silkie: Preference distillation for large visual language models, 2023.

\bibitem[Liu et~al.(2025)Liu, Sun, Zang, Dong, Cao, Duan, Lin, and Wang]{liu2025visual}
Ziyu Liu, Zeyi Sun, Yuhang Zang, Xiaoyi Dong, Yuhang Cao, Haodong Duan, Dahua Lin, and Jiaqi Wang.
\newblock Visual-rft: Visual reinforcement fine-tuning.
\newblock \emph{arXiv preprint arXiv:2503.01785}, 2025.

\bibitem[Bradley and Terry(1952)]{bradley1952rank}
Ralph~Allan Bradley and Milton~E Terry.
\newblock Rank analysis of incomplete block designs: I. the method of paired comparisons.
\newblock \emph{Biometrika}, 39\penalty0 (3/4):\penalty0 324--345, 1952.

\bibitem[Pi et~al.(2024{\natexlab{a}})Pi, Han, Xiong, Zhang, Liu, Pan, and Zhang]{pi2024strengthening}
Renjie Pi, Tianyang Han, Wei Xiong, Jipeng Zhang, Runtao Liu, Rui Pan, and Tong Zhang.
\newblock Strengthening multimodal large language model with bootstrapped preference optimization, 2024{\natexlab{a}}.

\bibitem[Yu et~al.(2024{\natexlab{a}})Yu, Zhang, Yao, Dang, Chen, Lu, Cui, He, Liu, Chua, et~al.]{yu2024rlaif}
Tianyu Yu, Haoye Zhang, Yuan Yao, Yunkai Dang, Da~Chen, Xiaoman Lu, Ganqu Cui, Taiwen He, Zhiyuan Liu, Tat-Seng Chua, et~al.
\newblock {RLAIF-V}: Aligning mllms through open-source ai feedback for super gpt-4v trustworthiness.
\newblock \emph{arXiv preprint arXiv:2405.17220}, 2024{\natexlab{a}}.

\bibitem[Rafailov et~al.(2023)Rafailov, Sharma, Mitchell, Ermon, Manning, and Finn]{DPO}
Rafael Rafailov, Archit Sharma, Eric Mitchell, Stefano Ermon, Christopher~D Manning, and Chelsea Finn.
\newblock Direct preference optimization: Your language model is secretly a reward model.
\newblock In \emph{NeurIPS}, 2023.

\bibitem[Yuan et~al.(2024)Yuan, Pang, Cho, Sukhbaatar, Xu, and Weston]{Self-Rewarding}
Weizhe Yuan, Richard~Yuanzhe Pang, Kyunghyun Cho, Sainbayar Sukhbaatar, Jing Xu, and Jason Weston.
\newblock Self-rewarding language models.
\newblock In \emph{ICML}, 2024.

\bibitem[Xiong et~al.(2024{\natexlab{a}})Xiong, Dong, Ye, Wang, Zhong, Ji, Jiang, and Zhang]{Iterative}
Wei Xiong, Hanze Dong, Chenlu Ye, Ziqi Wang, Han Zhong, Heng Ji, Nan Jiang, and Tong Zhang.
\newblock Iterative preference learning from human feedback: Bridging theory and practice for rlhf under kl-constraint.
\newblock In \emph{ICML}, 2024{\natexlab{a}}.

\bibitem[Lee et~al.(2023)Lee, Phatale, Mansoor, Mesnard, Ferret, Lu, Bishop, Hall, Carbune, Rastogi, et~al.]{lee2023rlaif}
Harrison Lee, Samrat Phatale, Hassan Mansoor, Thomas Mesnard, Johan Ferret, Kellie Lu, Colton Bishop, Ethan Hall, Victor Carbune, Abhinav Rastogi, et~al.
\newblock Rlaif vs. rlhf: Scaling reinforcement learning from human feedback with ai feedback.
\newblock \emph{arXiv preprint arXiv:2309.00267}, 2023.

\bibitem[Zelikman et~al.(2022)Zelikman, Wu, Mu, and Goodman]{Star}
Eric Zelikman, Yuhuai Wu, Jesse Mu, and Noah Goodman.
\newblock Star: Bootstrapping reasoning with reasoning.
\newblock In \emph{NeurIPS}, 2022.

\bibitem[Dong et~al.(2023)Dong, Xiong, Goyal, Zhang, Chow, Pan, Diao, Zhang, SHUM, and Zhang]{RAFT}
Hanze Dong, Wei Xiong, Deepanshu Goyal, Yihan Zhang, Winnie Chow, Rui Pan, Shizhe Diao, Jipeng Zhang, KaShun SHUM, and Tong Zhang.
\newblock {RAFT}: Reward ranked finetuning for generative foundation model alignment.
\newblock \emph{TMLR}, 2023.

\bibitem[Dong et~al.(2024)Dong, Xiong, Pang, Wang, Zhao, Zhou, Jiang, Sahoo, Xiong, and Zhang]{RLHF-Workflow}
Hanze Dong, Wei Xiong, Bo~Pang, Haoxiang Wang, Han Zhao, Yingbo Zhou, Nan Jiang, Doyen Sahoo, Caiming Xiong, and Tong Zhang.
\newblock Rlhf workflow: From reward modeling to online rlhf.
\newblock \emph{TMLR}, 2024.

\bibitem[Chen et~al.(2024{\natexlab{a}})Chen, Deng, Yuan, Ji, and Gu]{chen2024self}
Zixiang Chen, Yihe Deng, Huizhuo Yuan, Kaixuan Ji, and Quanquan Gu.
\newblock Self-play fine-tuning converts weak language models to strong language models.
\newblock \emph{arXiv preprint arXiv:2401.01335}, 2024{\natexlab{a}}.

\bibitem[Pi et~al.(2024{\natexlab{b}})Pi, Zhang, Zhang, Pan, Chen, and Zhang]{pi2024image}
Renjie Pi, Jianshu Zhang, Jipeng Zhang, Rui Pan, Zhekai Chen, and Tong Zhang.
\newblock Image textualization: An automatic framework for creating accurate and detailed image descriptions.
\newblock \emph{arXiv preprint arXiv:2406.07502}, 2024{\natexlab{b}}.

\bibitem[Chiu et~al.(2025)Chiu, Liu, Sapra, Tao, Jacoob, Ma, Yu, and Liu]{chiu2025aide}
Ming-Chang Chiu, Fuxiao Liu, Karan Sapra, Andrew Tao, Yaser Jacoob, Xuezhe Ma, Zhiding Yu, and Guilin Liu.
\newblock Aide: Agentically improve visual language model with domain experts.
\newblock \emph{arXiv preprint arXiv:2502.09051}, 2025.

\bibitem[Zhang et~al.(2024{\natexlab{a}})Zhang, Hosseini, Bansal, Kazemi, Kumar, and Agarwal]{zhang2024generative}
Lunjun Zhang, Arian Hosseini, Hritik Bansal, Mehran Kazemi, Aviral Kumar, and Rishabh Agarwal.
\newblock Generative verifiers: Reward modeling as next-token prediction.
\newblock \emph{arXiv preprint arXiv:2408.15240}, 2024{\natexlab{a}}.

\bibitem[Zhong et~al.(2024)Zhong, Feng, Xiong, Cheng, Zhao, He, Bian, and Wang]{zhong2024dpo}
Han Zhong, Guhao Feng, Wei Xiong, Xinle Cheng, Li~Zhao, Di~He, Jiang Bian, and Liwei Wang.
\newblock Dpo meets ppo: Reinforced token optimization for rlhf.
\newblock \emph{arXiv preprint arXiv:2404.18922}, 2024.

\bibitem[Yang et~al.(2024{\natexlab{a}})Yang, Ding, Lin, Zhang, and Zhang]{yang2024regularizing}
Rui Yang, Ruomeng Ding, Yong Lin, Huan Zhang, and Tong Zhang.
\newblock Regularizing hidden states enables learning generalizable reward model for llms.
\newblock \emph{arXiv preprint arXiv:2406.10216}, 2024{\natexlab{a}}.

\bibitem[Wang et~al.(2024{\natexlab{a}})Wang, Zheng, Chen, Liu, Dou, Huang, Shen, Jin, Zhou, Shi, et~al.]{wang2024secrets}
Binghai Wang, Rui Zheng, Lu~Chen, Yan Liu, Shihan Dou, Caishuang Huang, Wei Shen, Senjie Jin, Enyu Zhou, Chenyu Shi, et~al.
\newblock Secrets of rlhf in large language models part ii: Reward modeling.
\newblock \emph{arXiv preprint arXiv:2401.06080}, 2024{\natexlab{a}}.

\bibitem[Zhou et~al.(2024)Zhou, Cui, Rafailov, Finn, and Yao]{povid}
Yiyang Zhou, Chenhang Cui, Rafael Rafailov, Chelsea Finn, and Huaxiu Yao.
\newblock Aligning modalities in vision large language models via preference fine-tuning.
\newblock \emph{arXiv preprint arXiv:2402.11411}, 2024.

\bibitem[Wei et~al.(2022)Wei, Wang, Schuurmans, Bosma, Ichter, Xia, Chi, Le, and Zhou]{Chain-of-Thought-Prompting}
Jason Wei, Xuezhi Wang, Dale Schuurmans, Maarten Bosma, Brian Ichter, Fei Xia, Ed~H. Chi, Quoc~V. Le, and Denny Zhou.
\newblock Chain-of-thought prompting elicits reasoning in large language models.
\newblock In \emph{NeurIPS}, 2022.

\bibitem[Pang et~al.(2024)Pang, Yuan, He, Cho, Sukhbaatar, and Weston]{pang2024iterative}
Richard~Yuanzhe Pang, Weizhe Yuan, He~He, Kyunghyun Cho, Sainbayar Sukhbaatar, and Jason Weston.
\newblock Iterative reasoning preference optimization.
\newblock \emph{Advances in Neural Information Processing Systems}, 37:\penalty0 116617--116637, 2024.

\bibitem[Liu et~al.(2023{\natexlab{a}})Liu, Zeng, Ren, Li, Zhang, Yang, Li, Yang, Su, Zhu, and Zhang]{liu2023grounding}
Shilong Liu, Zhaoyang Zeng, Tianhe Ren, Feng Li, Hao Zhang, Jie Yang, Chunyuan Li, Jianwei Yang, Hang Su, Jun Zhu, and Lei Zhang.
\newblock Grounding dino: Marrying dino with grounded pre-training for open-set object detection, 2023{\natexlab{a}}.

\bibitem[Wu et~al.(2019)Wu, Kirillov, Massa, Lo, and Girshick]{wu2019detectron2}
Yuxin Wu, Alexander Kirillov, Francisco Massa, Wan-Yen Lo, and Ross Girshick.
\newblock Detectron2.
\newblock \url{https://github.com/facebookresearch/detectron2}, 2019.

\bibitem[Schulman et~al.(2017)Schulman, Wolski, Dhariwal, Radford, and Klimov]{PPO}
John Schulman, Filip Wolski, Prafulla Dhariwal, Alec Radford, and Oleg Klimov.
\newblock Proximal policy optimization algorithms.
\newblock \emph{arXiv:1707.06347}, 2017.

\bibitem[Ouyang et~al.(2022)Ouyang, Wu, Jiang, Almeida, Wainwright, Mishkin, Zhang, Agarwal, Slama, Ray, et~al.]{ouyang2022training}
Long Ouyang, Jeffrey Wu, Xu~Jiang, Diogo Almeida, Carroll Wainwright, Pamela Mishkin, Chong Zhang, Sandhini Agarwal, Katarina Slama, Alex Ray, et~al.
\newblock Training language models to follow instructions with human feedback.
\newblock \emph{Advances in Neural Information Processing Systems}, 35:\penalty0 27730--27744, 2022.

\bibitem[Touvron et~al.(2023)Touvron, Martin, Stone, Albert, Almahairi, Babaei, Bashlykov, Batra, Bhargava, Bhosale, et~al.]{touvron2023llama2}
Hugo Touvron, Louis Martin, Kevin Stone, Peter Albert, Amjad Almahairi, Yasmine Babaei, Nikolay Bashlykov, Soumya Batra, Prajjwal Bhargava, Shruti Bhosale, et~al.
\newblock Llama 2: Open foundation and fine-tuned chat models.
\newblock \emph{arXiv preprint arXiv:2307.09288}, 2023.

\bibitem[OpenAI(2024)]{2024gpt4o}
OpenAI.
\newblock Hello gpt-4o, 2024.
\newblock URL \url{https://openai.com/index/hello-gpt-4o/}.

\bibitem[Bai et~al.(2022)Bai, Jones, Ndousse, Askell, Chen, DasSarma, Drain, Fort, Ganguli, Henighan, et~al.]{bai2022training}
Yuntao Bai, Andy Jones, Kamal Ndousse, Amanda Askell, Anna Chen, Nova DasSarma, Dawn Drain, Stanislav Fort, Deep Ganguli, Tom Henighan, et~al.
\newblock Training a helpful and harmless assistant with reinforcement learning from human feedback.
\newblock \emph{arXiv preprint arXiv:2204.05862}, 2022.

\bibitem[Google(2023)]{gemini2023}
Google.
\newblock Gemini: A family of highly capable multimodal models, 2023.
\newblock URL \url{https://storage.googleapis.com/deepmind-media/gemini/gemini_1_report.pdf}.

\bibitem[Liu et~al.(2023{\natexlab{b}})Liu, Li, Wu, and Lee]{liu2023llava}
Haotian Liu, Chunyuan Li, Qingyang Wu, and Yong~Jae Lee.
\newblock Visual instruction tuning, 2023{\natexlab{b}}.

\bibitem[Dai et~al.(2023)Dai, Li, Li, Tiong, Zhao, Wang, Li, Fung, and Hoi]{dai2023instructblip}
Wenliang Dai, Junnan Li, Dongxu Li, Anthony Meng~Huat Tiong, Junqi Zhao, Weisheng Wang, Boyang Li, Pascale Fung, and Steven Hoi.
\newblock Instructblip: Towards general-purpose vision-language models with instruction tuning, 2023.

\bibitem[Zhang et~al.(2024{\natexlab{b}})Zhang, Zhang, Gu, Zhou, Lipka, Yang, and Sun]{zhang2024llavar}
Yanzhe Zhang, Ruiyi Zhang, Jiuxiang Gu, Yufan Zhou, Nedim Lipka, Diyi Yang, and Tong Sun.
\newblock Llavar: Enhanced visual instruction tuning for text-rich image understanding, 2024{\natexlab{b}}.

\bibitem[Chen et~al.(2023)Chen, Li, Dong, Zhang, He, Wang, Zhao, and Lin]{chen2023sharegpt4v}
Lin Chen, Jinsong Li, Xiaoyi Dong, Pan Zhang, Conghui He, Jiaqi Wang, Feng Zhao, and Dahua Lin.
\newblock Sharegpt4v: Improving large multi-modal models with better captions, 2023.

\bibitem[Li et~al.(2024{\natexlab{b}})Li, Zhang, Guo, Zhang, Li, Zhang, Zhang, Zhang, Li, Liu, et~al.]{li2024llavaov}
Bo~Li, Yuanhan Zhang, Dong Guo, Renrui Zhang, Feng Li, Hao Zhang, Kaichen Zhang, Peiyuan Zhang, Yanwei Li, Ziwei Liu, et~al.
\newblock Llava-onevision: Easy visual task transfer.
\newblock \emph{arXiv preprint arXiv:2408.03326}, 2024{\natexlab{b}}.

\bibitem[Girshick(2015)]{girshick2015fast}
Ross Girshick.
\newblock Fast r-cnn.
\newblock In \emph{Proceedings of the IEEE international conference on computer vision}, pages 1440--1448, 2015.

\bibitem[Yao et~al.(2021)Yao, Pi, Xu, Zhang, Li, and Zhang]{yao2021g}
Lewei Yao, Renjie Pi, Hang Xu, Wei Zhang, Zhenguo Li, and Tong Zhang.
\newblock G-detkd: towards general distillation framework for object detectors via contrastive and semantic-guided feature imitation.
\newblock In \emph{Proceedings of the IEEE/CVF international conference on computer vision}, pages 3591--3600, 2021.

\bibitem[Carion et~al.(2020)Carion, Massa, Synnaeve, Usunier, Kirillov, and Zagoruyko]{carion2020end}
Nicolas Carion, Francisco Massa, Gabriel Synnaeve, Nicolas Usunier, Alexander Kirillov, and Sergey Zagoruyko.
\newblock End-to-end object detection with transformers.
\newblock In \emph{Computer Vision--ECCV 2020: 16th European Conference, Glasgow, UK, August 23--28, 2020, Proceedings, Part I 16}, pages 213--229. Springer, 2020.

\bibitem[Kim et~al.(2022)Kim, Ka, Ahn, Joo, Chun, and Kim]{kim2022globallocal}
Doyeon Kim, Woonghyun Ka, Pyungwhan Ahn, Donggyu Joo, Sehwan Chun, and Junmo Kim.
\newblock Global-local path networks for monocular depth estimation with vertical cutdepth, 2022.

\bibitem[Yang et~al.(2024{\natexlab{b}})Yang, Kang, Huang, Xu, Feng, and Zhao]{yang2024depth}
Lihe Yang, Bingyi Kang, Zilong Huang, Xiaogang Xu, Jiashi Feng, and Hengshuang Zhao.
\newblock Depth anything: Unleashing the power of large-scale unlabeled data, 2024{\natexlab{b}}.

\bibitem[Gu et~al.(2021)Gu, Lin, Kuo, and Cui]{gu2021open}
Xiuye Gu, Tsung-Yi Lin, Weicheng Kuo, and Yin Cui.
\newblock Open-vocabulary object detection via vision and language knowledge distillation.
\newblock \emph{arXiv preprint arXiv:2104.13921}, 2021.

\bibitem[Yao et~al.(2024)Yao, Pi, Han, Liang, Xu, Zhang, Li, and Xu]{yao2024detclipv3}
Lewei Yao, Renjie Pi, Jianhua Han, Xiaodan Liang, Hang Xu, Wei Zhang, Zhenguo Li, and Dan Xu.
\newblock Detclipv3: Towards versatile generative open-vocabulary object detection, 2024.

\bibitem[Lightman et~al.(2023)Lightman, Kosaraju, Burda, Edwards, Baker, Lee, Leike, Schulman, Sutskever, and Cobbe]{lightman2023let}
Hunter Lightman, Vineet Kosaraju, Yuri Burda, Harrison Edwards, Bowen Baker, Teddy Lee, Jan Leike, John Schulman, Ilya Sutskever, and Karl Cobbe.
\newblock Let's verify step by step.
\newblock In \emph{The Twelfth International Conference on Learning Representations}, 2023.

\bibitem[Zhang et~al.(2024{\natexlab{c}})Zhang, Wang, Diao, Lin, Pan, Dong, Zhang, Molchanov, and Zhang]{zhang2024entropy}
Hanning Zhang, Pengcheng Wang, Shizhe Diao, Yong Lin, Rui Pan, Hanze Dong, Dylan Zhang, Pavlo Molchanov, and Tong Zhang.
\newblock Entropy-regularized process reward model.
\newblock \emph{arXiv preprint arXiv:2412.11006}, 2024{\natexlab{c}}.

\bibitem[Wang et~al.(2023)Wang, Li, Shao, Xu, Dai, Li, Chen, Wu, and Sui]{wang2023math}
Peiyi Wang, Lei Li, Zhihong Shao, RX~Xu, Damai Dai, Yifei Li, Deli Chen, Yu~Wu, and Zhifang Sui.
\newblock Math-shepherd: Verify and reinforce llms step-by-step without human annotations.
\newblock \emph{arXiv preprint arXiv:2312.08935}, 2023.

\bibitem[Zheng et~al.(2023)Zheng, Chiang, Sheng, Zhuang, Wu, Zhuang, Lin, Li, Li, Xing, et~al.]{zheng2023judging}
Lianmin Zheng, Wei-Lin Chiang, Ying Sheng, Siyuan Zhuang, Zhanghao Wu, Yonghao Zhuang, Zi~Lin, Zhuohan Li, Dacheng Li, Eric Xing, et~al.
\newblock Judging llm-as-a-judge with mt-bench and chatbot arena.
\newblock \emph{Advances in Neural Information Processing Systems}, 36:\penalty0 46595--46623, 2023.

\bibitem[Xiong et~al.(2024{\natexlab{b}})Xiong, Wang, Guo, Ye, Fan, Gu, Huang, and Li]{xiong2024llavacritic}
Tianyi Xiong, Xiyao Wang, Dong Guo, Qinghao Ye, Haoqi Fan, Quanquan Gu, Heng Huang, and Chunyuan Li.
\newblock Llava-critic: Learning to evaluate multimodal models.
\newblock \emph{arXiv preprint arXiv:2410.02712}, 2024{\natexlab{b}}.

\bibitem[Xu et~al.(2023)Xu, Lee, Sukhbaatar, and Weston]{xu2023some}
Jing Xu, Andrew Lee, Sainbayar Sukhbaatar, and Jason Weston.
\newblock Some things are more cringe than others: Iterative preference optimization with the pairwise cringe loss.
\newblock \emph{arXiv preprint arXiv:2312.16682}, 2023.

\bibitem[Gulcehre et~al.(2024)Gulcehre, Paine, Srinivasan, Konyushkova, Weerts, Sharma, Siddhant, Ahern, Wang, Gu, et~al.]{ReST}
Caglar Gulcehre, Tom~Le Paine, Srivatsan Srinivasan, Ksenia Konyushkova, Lotte Weerts, Abhishek Sharma, Aditya Siddhant, Alex Ahern, Miaosen Wang, Chenjie Gu, et~al.
\newblock Reinforced self-training (rest) for language modeling.
\newblock In \emph{EMNLP}, 2024.

\bibitem[Chen et~al.(2024{\natexlab{b}})Chen, Deng, Yuan, Ji, and Gu]{chen2024selfplay}
Zixiang Chen, Yihe Deng, Huizhuo Yuan, Kaixuan Ji, and Quanquan Gu.
\newblock Self-play fine-tuning converts weak language models to strong language models, 2024{\natexlab{b}}.

\bibitem[Hu et~al.(2022)Hu, Shen, Wallis, Allen-Zhu, Li, Wang, Wang, Chen, et~al.]{hu2022lora}
Edward~J Hu, Yelong Shen, Phillip Wallis, Zeyuan Allen-Zhu, Yuanzhi Li, Shean Wang, Lu~Wang, Weizhu Chen, et~al.
\newblock Lora: Low-rank adaptation of large language models.
\newblock \emph{ICLR}, 1\penalty0 (2):\penalty0 3, 2022.

\bibitem[Lu et~al.(2024{\natexlab{a}})Lu, Jiang, Chen, Wang, Choi, and Lin]{lu2024wildvision}
Yujie Lu, Dongfu Jiang, Wenhu Chen, William~Yang Wang, Yejin Choi, and Bill~Yuchen Lin.
\newblock Wildvision: Evaluating vision-language models in the wild with human preferences.
\newblock \emph{arXiv preprint arXiv:2406.11069}, 2024{\natexlab{a}}.

\bibitem[Yu et~al.(2024{\natexlab{b}})Yu, Yao, Zhang, He, Han, Cui, Hu, Liu, Zheng, Sun, et~al.]{yu2024rlhfv}
Tianyu Yu, Yuan Yao, Haoye Zhang, Taiwen He, Yifeng Han, Ganqu Cui, Jinyi Hu, Zhiyuan Liu, Hai-Tao Zheng, Maosong Sun, et~al.
\newblock {RlHF-V}: Towards trustworthy mllms via behavior alignment from fine-grained correctional human feedback.
\newblock In \emph{CVPR}, 2024{\natexlab{b}}.

\bibitem[Yue et~al.(2024)Yue, Zheng, Ni, Wang, Zhang, Tong, Sun, Yu, Zhang, Sun, et~al.]{yue2024mmmu}
Xiang Yue, Tianyu Zheng, Yuansheng Ni, Yubo Wang, Kai Zhang, Shengbang Tong, Yuxuan Sun, Botao Yu, Ge~Zhang, Huan Sun, et~al.
\newblock Mmmu-pro: A more robust multi-discipline multimodal understanding benchmark.
\newblock \emph{arXiv preprint arXiv:2409.02813}, 2024.

\bibitem[Zhang et~al.(2024{\natexlab{d}})Zhang, Jiang, Zhang, Lin, Guo, Qiu, Zhou, Lu, Chang, Qiao, et~al.]{zhang2024mathverse}
Renrui Zhang, Dongzhi Jiang, Yichi Zhang, Haokun Lin, Ziyu Guo, Pengshuo Qiu, Aojun Zhou, Pan Lu, Kai-Wei Chang, Yu~Qiao, et~al.
\newblock Mathverse: Does your multi-modal llm truly see the diagrams in visual math problems?
\newblock In \emph{European Conference on Computer Vision}, pages 169--186. Springer, 2024{\natexlab{d}}.

\bibitem[Liang et~al.(2024)Liang, Liu, Niu, Zhang, Zhou, and Yavuz]{liang2024improving}
Zhenwen Liang, Ye~Liu, Tong Niu, Xiangliang Zhang, Yingbo Zhou, and Semih Yavuz.
\newblock Improving llm reasoning through scaling inference computation with collaborative verification.
\newblock \emph{arXiv preprint arXiv:2410.05318}, 2024.

\bibitem[Hosseini et~al.(2024)Hosseini, Yuan, Malkin, Courville, Sordoni, and Agarwal]{hosseini2024v}
Arian Hosseini, Xingdi Yuan, Nikolay Malkin, Aaron Courville, Alessandro Sordoni, and Rishabh Agarwal.
\newblock V-star: Training verifiers for self-taught reasoners.
\newblock \emph{arXiv preprint arXiv:2402.06457}, 2024.

\bibitem[Wang et~al.(2024{\natexlab{b}})Wang, Bai, Tan, Wang, Fan, Bai, Chen, Liu, Wang, Ge, et~al.]{wang2024qwen2}
Peng Wang, Shuai Bai, Sinan Tan, Shijie Wang, Zhihao Fan, Jinze Bai, Keqin Chen, Xuejing Liu, Jialin Wang, Wenbin Ge, et~al.
\newblock Qwen2-vl: Enhancing vision-language model's perception of the world at any resolution.
\newblock \emph{arXiv preprint arXiv:2409.12191}, 2024{\natexlab{b}}.

\bibitem[Chen et~al.(2024{\natexlab{c}})Chen, Wang, Cao, Liu, Gao, Cui, Zhu, Ye, Tian, Liu, et~al.]{chen2024expanding}
Zhe Chen, Weiyun Wang, Yue Cao, Yangzhou Liu, Zhangwei Gao, Erfei Cui, Jinguo Zhu, Shenglong Ye, Hao Tian, Zhaoyang Liu, et~al.
\newblock Expanding performance boundaries of open-source multimodal models with model, data, and test-time scaling.
\newblock \emph{arXiv preprint arXiv:2412.05271}, 2024{\natexlab{c}}.

\bibitem[Liu et~al.(2024)Liu, Li, Li, Li, Zhang, Shen, and Lee]{liu2024llavanext}
Haotian Liu, Chunyuan Li, Yuheng Li, Bo~Li, Yuanhan Zhang, Sheng Shen, and Yong~Jae Lee.
\newblock Llava-next: Improved reasoning, ocr, and world knowledge, January 2024.
\newblock URL \url{https://llava-vl.github.io/blog/2024-01-30-llava-next/}.

\bibitem[Bai et~al.(2023)Bai, Bai, Yang, Wang, Tan, Wang, Lin, Zhou, and Zhou]{bai2023qwenvl}
Jinze Bai, Shuai Bai, Shusheng Yang, Shijie Wang, Sinan Tan, Peng Wang, Junyang Lin, Chang Zhou, and Jingren Zhou.
\newblock Qwen-vl: A versatile vision-language model for understanding, localization, text reading, and beyond, 2023.

\bibitem[Dubey et~al.(2024)Dubey, Jauhri, Pandey, Kadian, Al-Dahle, Letman, Mathur, Schelten, Yang, Fan, et~al.]{dubey2024llama}
Abhimanyu Dubey, Abhinav Jauhri, Abhinav Pandey, Abhishek Kadian, Ahmad Al-Dahle, Aiesha Letman, Akhil Mathur, Alan Schelten, Amy Yang, Angela Fan, et~al.
\newblock The llama 3 herd of models.
\newblock \emph{arXiv preprint arXiv:2407.21783}, 2024.

\bibitem[Deitke et~al.(2024)Deitke, Clark, Lee, Tripathi, Yang, Park, Salehi, Muennighoff, Lo, Soldaini, et~al.]{deitke2024molmo}
Matt Deitke, Christopher Clark, Sangho Lee, Rohun Tripathi, Yue Yang, Jae~Sung Park, Mohammadreza Salehi, Niklas Muennighoff, Kyle Lo, Luca Soldaini, et~al.
\newblock Molmo and pixmo: Open weights and open data for state-of-the-art multimodal models.
\newblock \emph{arXiv preprint arXiv:2409.17146}, 2024.

\bibitem[Lu et~al.(2024{\natexlab{b}})Lu, Liu, Zhang, Wang, Dong, Liu, Sun, Ren, Li, Yang, et~al.]{lu2024deepseek}
Haoyu Lu, Wen Liu, Bo~Zhang, Bingxuan Wang, Kai Dong, Bo~Liu, Jingxiang Sun, Tongzheng Ren, Zhuoshu Li, Hao Yang, et~al.
\newblock Deepseek-vl: towards real-world vision-language understanding.
\newblock \emph{arXiv preprint arXiv:2403.05525}, 2024{\natexlab{b}}.

\bibitem[Li et~al.(2024{\natexlab{c}})Li, Liu, Wu, Wang, Shen, Qu, Niu, Wang, Chen, and Li]{li2024aria}
Dongxu Li, Yudong Liu, Haoning Wu, Yue Wang, Zhiqi Shen, Bowen Qu, Xinyao Niu, Guoyin Wang, Bei Chen, and Junnan Li.
\newblock Aria: An open multimodal native mixture-of-experts model.
\newblock \emph{arXiv preprint arXiv:2410.05993}, 2024{\natexlab{c}}.

\bibitem[Guo et~al.(2024)Guo, Zheng, Bai, Li, Wang, Zhu, Li, Neubig, Chen, and Yue]{guo2024mammoth}
Jarvis Guo, Tuney Zheng, Yuelin Bai, Bo~Li, Yubo Wang, King Zhu, Yizhi Li, Graham Neubig, Wenhu Chen, and Xiang Yue.
\newblock Mammoth-vl: Eliciting multimodal reasoning with instruction tuning at scale.
\newblock \emph{arXiv preprint arXiv:2412.05237}, 2024.

\bibitem[Yang et~al.(2024{\natexlab{c}})Yang, Yang, Zhang, Hui, Zheng, Yu, Li, Liu, Huang, Wei, et~al.]{yang2024qwen2}
An~Yang, Baosong Yang, Beichen Zhang, Binyuan Hui, Bo~Zheng, Bowen Yu, Chengyuan Li, Dayiheng Liu, Fei Huang, Haoran Wei, et~al.
\newblock Qwen2. 5 technical report.
\newblock \emph{arXiv preprint arXiv:2412.15115}, 2024{\natexlab{c}}.

\end{thebibliography}

\clearpage
\setcounter{page}{1}
\setcounter{section}{0}

\appendix

\section{Formalulation of RM Training Baselines}
\label{appendix:formulation}

\noindent \textbf{Supervised Fine-Tuning (SFT)} Supervised fine-tuning (SFT) is a standard method for training LLMs, optimizing the cross-entropy loss between the predicted and actual target tokens. Given a dataset \( \mathcal{D} = \{ (\mathbf{x}, y) \} \) of input context \( \mathbf{x} \) and target response \( y \), the SFT objective is:

\begin{equation}
    \mathcal{L}_{\text{SFT}}(\theta, \mathcal{D}) = 
    - \mathbb{E}_{( \mathbf{x}, y) \sim \mathcal{D}} 
    \left[
    \sum_{t=1}^{|y|} \log p_{\theta}(y_t \mid \mathbf{x}, y_{<t})
    \right].
\end{equation}

\noindent \textbf{Best-of-N (BoN)} Given an image \( I \) and input context \( \mathbf{x} \), BoN generates \( N \) candidate responses from a vision-language model and selects the highest-scoring one using a reward model or verifier:

\begin{equation}
\begin{aligned}
y_{\text{BoN}}(I, \mathbf{x}) = \arg\max_{y \in \mathcal{Y}_{\text{gen}}} r_{\theta}(I, \mathbf{x}, y).
\end{aligned}
\end{equation}

BoN improves inference by ranking multiple responses and selecting the most likely correct one, enhancing reasoning accuracy in tasks like VQA and diagram-based problem solving.

\noindent \textbf{Direct Preference Optimization (DPO).}  
DPO simplifies policy optimization by reformulating Proximal Policy Optimization (PPO) using BT model, enabling direct alignment with human preferences. Given an image \( I \), input context \( \mathbf{x} \), and preference pairs \( (y^+, y^-) \), the objective is:

\begin{equation}
\begin{aligned}
& \mathcal{L}_{\text{DPO}}(\theta) = \\
& -\mathbb{E}_{(I, \mathbf{x}, y^+, y^-) \sim \mathcal{D}} 
\left[ 
\log \sigma \Big( q_{\theta}(I, \mathbf{x}, y^+) - q_{\theta}(I, \mathbf{x}, y^-) \Big) 
\right],
\end{aligned}
\end{equation}

where \( q_{\theta}(I, \mathbf{x}, y) \) is the reward-aligned log probability.  
DPO eliminates the need for explicit reward modeling in RLHF, but it is also frequently used for training reward models themselves.

\noindent \textbf{VLM-as-a-Judge.} Instead of fine-tuning a dedicated verifier, VLM-as-a-Judge directly prompts a vision-language model (VLM) to evaluate candidate responses \( y \) given an image \( I \) and input context \( \mathbf{x} \). The model assigns a score based on its assessment:

\[
r_{\theta}(I, \mathbf{x}, y) = f_{\theta}(I, \mathbf{x}, y, \mathbf{p}),
\]

where \( f_{\theta} \) represents the VLM's scoring function, and \( \mathbf{p} \) is an optional prompt guiding the evaluation.  
VLM-as-a-Judge can also perform \textit{reference-guided grading}, comparing \( y \) with a reference solution \( y^* \) to provide a more structured evaluation.

\section{Implementation}
\label{appendix:implementation}




All training is conducted on 8 A6000 GPUs. We fine-tune the full model for Rewarding Instruction-Following Fine-Tuning and apply LoRA for Iterative Optimization. For Cold-Start Pairwise Data Generation, we extract 10K samples from the combination of ShareGPT-V~\citep{chen2023sharegpt4v}, LLaVAR~\citep{zhang2024llavar} and LLaVA-Instruct~\citep{liu2023llava} datasets and use GroundingDINO~\citep{liu2023grounding} and Detectron2~\citep{wu2019detectron2} to verify object presence, refining rejected responses. Qwen2.5-32B-Instruct~\citep{yang2024qwen2} is used to generate replacements for "approximately correct" responses. In CoT Generation and Verification, we employ Qwen2-VL-72B-Instruct~\citep{wang2024qwen2} to generate CoT critiques, incorporating detected objects to reduce hallucinations. The dataset is formatted into a two-round dialogue with a Yes/No reward signal. During Iterative Training, we extract 5K additional samples per iteration, and fine-tune the model using LoRA to enhance reward consistency. 

\subsection{Training Hyperparameters}
During training, we employ a warm-up ratio of 0.1, a global batch size of 32, and a learning rate of 2.0e-5, a epoch num of 1. For Qwen, we use a cutoff length of 2048, while for InternVL, we adopt a cutoff length of 4096. In iterative optimization, we apply a LoRA rank of 16 with a learning rate of 2e-4.

\subsection{Other Hyperparameters}
During iterative training, iteration 1 uses a margin of [0.3, 1], while iteration 2 uses a margin of [0.3, 0.99]. For best-of-n evaluation, we employ a temperature of 0.7, a top-p value of 0.9, and a top-k value of 50.

\begin{wrapfigure}{r}{0.48\textwidth}
\vspace{-1.8cm}
  \centering
    \includegraphics[width=\linewidth]{ 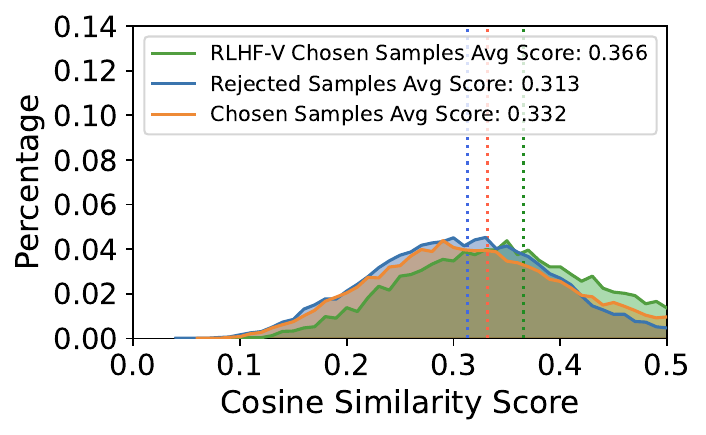}

   \caption{Contamination study of the dataset used in this work. The results show that the improvement is not merely due to including data from the same distribution.
   }
   \label{fig:contamination}
\end{wrapfigure}
\section{Data Contamination Analysis}
\label{appendix:contamination}

We concatenate the prompt and the response of one sample into a single text and apply TF-IDF vectorization. For each sample, we compute the cosine similarity between its resulting vector and the vectors of all samples in VLRewardBench. The highest similarity score among them is taken as the similarity of the sample. By analyzing the similarity distribution across samples, we assess data contamination. RLHF-V is used as a baseline for comparison in this analysis. The results in Figure \ref{fig:contamination} show that the improvement is not merely due to including data from the same distribution.

\section{Full Results for VLM comparison}
\label{appendix:full_vlm_comparison}
\begin{table}[ht]
    \centering
    \resizebox{\textwidth}{!}{
    \begin{tabular}{l c c c c}
        \toprule
        \textbf{Model} & \textbf{General} & \textbf{Hallu.} & \textbf{Reasoning} & \textbf{Overall} \\
        \midrule
        \multicolumn{5}{l}{\textit{Proprietary VLMs}} \\
        GPT-4o & 49.1 & 67.6 & 70.5 & 65.8 \\
        Gemini-1.5-Pro & 50.8 & 72.5 & 64.2 & 67.2 \\
        Claude-3.5-Sonnet & 43.4 & 55.0 & 62.3 & 55.3 \\
        GPT-4o-mini & 41.7& 34.5 & 58.2	& 41.5 \\
        Qwen-VL-Max & 40.6 & 46.0 & 57.6 & 48.2 \\
        \midrule
        \multicolumn{5}{l}{\textit{Open-Source VLMs}} \\
        Llama-3.2-90B & 42.6 & 57.3 & 61.7 & 56.2 \\
        Qwen2-VL-72B & 38.1 & 32.8 & 58.0 & 39.5 \\
        Molmo-72B & 33.9 & 42.3 & 54.9 & 44.1 \\
        NVLM-D-72B & 38.9 & 31.6 & 62.0 & 40.1 \\
        QvQ-72B & 41.8 &46.2 &51.2 &46.4 \\
        DeepSeek-VL-27B & 29.7 & 23.8 & 50.9 & 30.3 \\
        Aria-25B & 37.9 & 33.1 & 64.0 & 41.1 \\
        Pixtral-12B & 35.6 & 25.9 & 59.9 & 35.8 \\
        Llama-3.2-11B & 33.3 & 38.4 & 56.6 & 42.9 \\
        InternVL2-8B & 35.6 & 41.1 & 59.0 & 44.5 \\
        MAmmoTH-VL-8B & 42.0 & 41.0 & 53.0 & 45.2 \\
        Qwen2-VL-7B & 31.6 & 19.1 & 51.1 & 28.3 \\
        LLaVA-OneVision-7B-ov & 32.2 & 20.1 & 57.1 & 29.6 \\
        Molmo-7B & 31.1 & 31.8 & 56.2 & 37.5 \\
        Phi-3.5-Vision-4B & 28.0 & 22.4 & 56.6 & 28.2 \\
        \midrule
        VL-GenRM & 51.9 & 81.4 & 54.4 & 70.2 \\
        \bottomrule
    \end{tabular}}
    \vspace{0.2cm}
    \caption{Comparison of Proprietary and Open-Source VLMs on VLRewardBench datasets. We observe consistent and significant performance boosts over the comparison models on general and hallucination parts.}
    \label{tab:vlm_main_comparison}
\end{table}

\section{Data Examples}
\label{appendix:data_examples}

We provide two examples of the CoT generated by the two models during iteration in Table \ref{tab:exp1} and Table \ref{tab:exp2}.

\begin{table*}[t]\centering
\begin{minipage}{1.0\textwidth}\vspace{0mm}    \centering
\begin{sectionbox}[]{Example1}
    \centering
      \footnotesize
\begin{tabular}{p{0.97\textwidth} c}

\includegraphics[width=0.4\linewidth]{ }

[Prompt] \\
Describe the following image. \\

\vspace{1pt}
[chosen Response] \\
The image presents a striking scene of a {\color{blue}zebra} in its natural habitat. The {\color{blue}zebra}, with its distinctive {\color{blue}black and white stripes}, stands as the central figure in this composition. It is facing to the right, perhaps gazing into the distance or observing its surroundings. \\

The setting is a {\color{blue}field, which appears dry and brown}, indicative of a harsh environment or perhaps a particular season. Despite the arid conditions, there are {\color{blue}patches of green grass} that provide a stark contrast to the otherwise monotonous landscape. \\

Scattered throughout the field are {\color{blue}rocks} of various sizes. Their random placement adds an element of ruggedness to the scene. \\

The {\color{blue}zebra} is positioned near the center of the image, drawing the viewer's attention immediately. Its location, combined with its contrasting stripes against the muted background, makes it the undeniable focal point of this picture. \\ 

Overall, this image captures a moment of calm in the wild, with the {\color{blue}zebra} in its element amidst a rugged landscape. \\
\vspace{1pt}

[CoT] \\
The response accurately describes the {\color{blue}zebra} and its surroundings, including the {\color{blue}dry field, green patches, and scattered rocks}. However, it does not mention the zebra's {\color{blue}tail}, which is visible in the image and is a notable feature of the animal. \\
\vspace{1pt}

[Rejected Response] \\
The image presents a striking scene of a deer in its natural habitat. The {\color{red}deer}, with its distinctive brown and white spots, stands as the central figure in this composition. It is facing to the right, perhaps gazing into the distance or observing its surroundings. \\

The setting is a field, which appears dry and brown, indicative of a harsh environment or perhaps a particular season. Despite the arid conditions, there are patches of green grass that provide a stark contrast to the otherwise monotonous landscape. \\

Scattered throughout the field are rocks of various sizes. Their random placement adds an element of ruggedness to the scene. \\

The deer is positioned near the center of the image, drawing the viewer's attention immediately. Its location, combined with its contrasting spots against the muted background, makes it the undeniable focal point of this picture. \\

Overall, this image captures a moment of calm in the wild, with the deer in its element amidst a rugged landscape. \\
\vspace{1pt}

[CoT] \\
The response incorrectly identifies the animal as a {\color{red}deer}. The animal in the image is actually a {\color{red}zebra, which is characterized by its distinctive black and white stripes}.

\end{tabular}
\end{sectionbox}
\vspace{-2mm}
\caption{One example of generated CoT during iteration.}
    \label{tab:exp1}
\end{minipage}
\end{table*}

\begin{table*}[t]\centering
\begin{minipage}{1.0\textwidth}\vspace{0mm}    \centering
\begin{sectionbox}[]{Example2}
    \centering
      \footnotesize
\begin{tabular}{p{0.97\textwidth} c}

\includegraphics[width=0.4\linewidth]{ 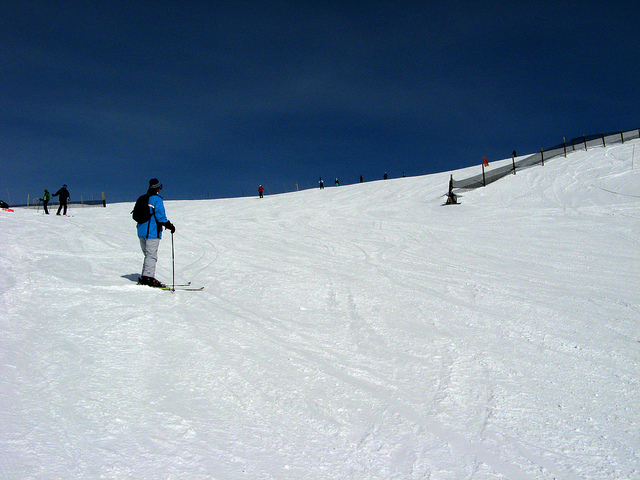}

[Prompt] \\
What sport is this onlooker attempting? \\

\vspace{1pt}
[chosen Response] \\
Skiing \\
\vspace{1pt}

[CoT] \\
The response is correct. The sport being attempted is skiing, as evidenced by the {\color{blue}skier's equipment and the snowy environment}. \\
\vspace{1pt}

[Rejected Response] \\
Basketball \\
\vspace{1pt}

[CoT] \\
The response incorrectly identifies the sport as {\color{red}basketball}. The image clearly shows a person skiing down a snowy slope, which is a key activity in the sport of skiing. {\color{red}The presence of skis, ski poles, and the snowy environment are indicative of skiing, not basketball}.

\end{tabular}
\end{sectionbox}
\vspace{-2mm}
\caption{One example of generated CoT during iteration.}
    \label{tab:exp2}
\end{minipage}
\end{table*}

\clearpage
\section{Prompt Template}
\label{appendix:prompts}

\begin{table*}[t]\centering
\begin{minipage}{1.0\textwidth}\vspace{0mm}    \centering
\begin{sectionbox}[]{System prompt for object extraction}
    \centering
      \footnotesize
\begin{tabular}{p{0.97\textwidth} c}

\begin{lstlisting}[breaklines=True]
###TASK DESCRIPTION###
Your are a helpful entities extractor. Please help me to extract the OBJECTS mentioned in a description about an image. 
###ATTENTION###
1. Only extract the descriptions of objects that are described with certainty. For example, in the sentence "there's a white car parked, perhaps belonging to one of the hotel guests," the "hotel guests" part is included within "perhaps," indicating uncertainty. Therefore, you only need extract "a white car" that is described with certainty. 
2. Avoid extracting abstract or non-specific entities (such as "cozy atmosphere", "excitement", "sky view")!!! 
3. Your response should strictly start with "%%%RESPONSE%%%:" and follow this format: "%%%RESPONSE%%%: obj1. obj2. obj3...."

###IN-CONTEXT EXAMPLES###
Here are some examples for your reference.

Example1: {Example1}

Example2: {Example2}

Example3: {Example3}
\end{lstlisting}

\end{tabular}
\end{sectionbox}
\vspace{-2mm}
\caption{System prompt for object extraction. The content within \{\} will be replaced with the corresponding values.}
    \label{tab:prompt}
\end{minipage}
\end{table*}

\begin{table*}[t]\centering
\begin{minipage}{1.0\textwidth}\vspace{0mm}    \centering
\begin{sectionbox}[]{System prompt for identifying false rejections.}
    \centering
      \footnotesize
\begin{tabular}{p{0.97\textwidth} c}

\begin{lstlisting}[breaklines=True]
You are an AI model designed to evaluate the accuracy of responses based on a given image, a list of detected objects from the image, and a provided prompt. Your task is to determine whether the response is generally correct in relation to the image, considering the detected objects as a reference.  

### **Evaluation Criteria:**  
1. **General Accuracy:**  
   - The response should be broadly correct based on the image content.  
   - Minor inaccuracies or missing details are acceptable as long as the overall meaning remains correct.  

2. **Consistency with Detected Objects:**  
   - The detected object list may not include all objects in the image.  
   - If the response describes something that is in the detected object list, it is likely correct.  
   - If the response describes something not in the detected object list, it may still be correct if it is a reasonable inference.  

3. **Reasonable Inference:**  
   - The response can include reasonable assumptions based on the detected objects.  
   - However, if the response describes something that contradicts the detected objects or is highly specific without clear evidence, it should be considered incorrect.  

4. **Handling Uncertainty:**  
   - If the response makes a claim that cannot be verified based on the detected objects alone, but is plausible, consider it correct.  
   - If the response makes a claim that is clearly false based on the detected objects, consider it incorrect.  

### **Output Format:**  
Respond with only one word:  
- Correct: If the response is generally accurate.  
- Incorrect: If the response contains clear errors or contradictions.
\end{lstlisting}

\end{tabular}
\end{sectionbox}
\vspace{-2mm}
\caption{System prompt for identifying false rejections.}
    \label{tab:prompt}
\end{minipage}
\end{table*}

\begin{table*}[t]\centering
\begin{minipage}{1.0\textwidth}\vspace{0mm}    \centering
\begin{sectionbox}[]{Prompt template for altering objects}
    \centering
      \footnotesize
\begin{tabular}{p{0.97\textwidth} c}

\begin{lstlisting}[breaklines=True]
I will give you a description of a image and some objects in the description, and your task is to replace these objects with any other objects. Please output the modified description.

Description: {Description}

Objects: {Objects}
\end{lstlisting}

\end{tabular}
\end{sectionbox}
\vspace{-2mm}
\caption{Prompt template for altering objects. The content within \{\} will be replaced with the corresponding values.}
    \label{tab:prompt}
\end{minipage}
\end{table*}

\begin{table*}[t]\centering
\begin{minipage}{1.0\textwidth}\vspace{0mm}    \centering
\begin{sectionbox}[]{System prompt for generating critique CoT }
    \centering
      \footnotesize
\begin{tabular}{p{0.97\textwidth} c}

\begin{lstlisting}[breaklines=True]
You are an advanced Multimodal Large Language Model. Your task is to generate a critique of the response based on an image, a prompt, and an extracted object list.  

### **Input**
1. **Image**
2. **Object List**: The object list contains some, but not all, objects in the image. It can help identify missing elements, but you must **not rely solely on it** always refer to the image itself.
3. **Prompt**
4. **Response**

### **Output**
**Critique**

### **Note**
1. Critiques should include only **crucial and obvious** errors in the MLLM's response and provide clear justifications for why they are incorrect.
2. If an object is **semantically important** to the scene or context, its absence in the response is considered an error. However, the omission of **minor or background objects** is acceptable unless they are explicitly relevant to the prompt.
3. Please only point out the errors and do not judge whether the response is correct or not.
\end{lstlisting}

\end{tabular}
\end{sectionbox}
\vspace{-2mm}
\caption{System prompt for generating critique CoT.}
    \label{tab:critique_prompt}
\end{minipage}
\end{table*}

\begin{table*}[t]\centering
\begin{minipage}{1.0\textwidth}\vspace{0mm}    \centering
\begin{sectionbox}[]{System prompt for generating descriptive CoT rationales}
    \centering
      \footnotesize
\begin{tabular}{p{0.97\textwidth} c}

\begin{lstlisting}[breaklines=True]
You are an AI assistant designed to evaluate the quality of responses.

### **Input**
1. **Image**
2. **Object List**: The object list contains some, but not all, objects in the image. It can help identify missing elements, but you must **not rely solely on it** always refer to the image itself.
3. **Prompt**
4. **Response**
5. **Judgment**: Yes or No, indicating whether the response is of high quality

### **Output**
**Explanation**: A step-by-step Chain-of-Thought (CoT) explanation that justifies why the response is considered high quality (Yes) or not (No).
\end{lstlisting}

\end{tabular}
\end{sectionbox}
\vspace{-2mm}
\caption{System prompt for generating descriptive CoT rationales.}
    \label{tab:descriptive_prompt}
\end{minipage}
\end{table*}

\begin{table*}[t]\centering
\begin{minipage}{1.0\textwidth}\vspace{0mm}    \centering
\begin{sectionbox}[]{Prompt template for training.}
    \centering
      \footnotesize
\begin{tabular}{p{0.97\textwidth} c}

\textbf{LLM as a judge}

\begin{lstlisting}[breaklines=True]
[User]
<image>Given an image and a corresponding prompt, please serve as an unbiased and fair judge to evaluate the quality of the response provided by a Large Multimodal Model (LMM). Determine which response is better and explain your reasoning with specific details. Your task is provided as follows:
Prompt: {0}
Response A: {Response A}
Response B: {Response B}
\end{lstlisting}

\\
\textbf{GenRM}

\begin{lstlisting}[breaklines=True]
[User]
<image>Given an image and a corresponding prompt, please serve as an unbiased and fair judge to evaluate the quality of the response provided by a Large Multimodal Model (LMM). Your task is provided as follows:
Prompt: {Prompt}
Response: {Response}

[Assistant]
{Yes / No}
\end{lstlisting}

\\
\textbf{IFT}

\begin{lstlisting}[breaklines=True]
[User]
<image>Given an image and a corresponding prompt, please serve as an unbiased and fair judge to evaluate the quality of the response provided by a Large Multimodal Model (LMM). Your task is provided as follows:
Prompt: {Prompt}
Response: {Response}
{Generate a critique of the response first. / Let's think step by step}

[Assistant]
{CoT}

[User]
Overall, is this response of high quality?

[Assistant]
{Yes / No}
\end{lstlisting}

\end{tabular}
\end{sectionbox}
\vspace{-2mm}
\caption{Prompt template for training. The content within \{\} will be replaced with the corresponding values.}
    \label{tab:prompt}
\end{minipage}
\end{table*}

\end{document}